%% file: MAIN.tex
\PassOptionsToPackage{table,xcdraw}{xcolor}
\documentclass[acmsmall,natbib=false]{acmart}

\setcopyright{none}

\AtBeginDocument{%
  }

\usepackage[algo2e]{algorithm2e}
\usepackage{graphicx}
\usepackage{textcomp}
\usepackage[table,xcdraw]{xcolor}
\usepackage{soul}
\usepackage{array}
\usepackage{listings}
\usepackage{xspace}
\usepackage{hyperref}
\usepackage{multirow}
\usepackage{algorithm}
\usepackage{algorithmic}
\usepackage{amsmath, amsfonts}
\usepackage{multirow}
\usepackage{float}
\usepackage{wrapfig,lipsum,booktabs}
\usepackage{subcaption}
\usepackage{color}
\usepackage{tabularray}
\usepackage{balance}
\usepackage[most, breakable]{tcolorbox}
\usepackage{hyperref}
\usepackage{url}

\newcommand{\dataset}{VERT\xspace}

\newcommand{\ie}{\textit{i.e.,}\xspace}
\newcommand{\eg}{\textit{e.g.,}\xspace}

\newtcolorbox[auto counter, number within=section]{mybox}[2][]{%
    title=\thetcbcounter: #2, #1, breakable} 

\newtcolorbox{fncyBox}{
enhanced,
boxrule=0pt, frame hidden,
borderline west={4pt}{0pt}{green!75!black},
colback=green!10!white,
sharp corners
}

\RequirePackage[
  datamodel=acmdatamodel,
  style=acmauthoryear,
  ]{biblatex}

\begin{document}

\title{Enhancing Large Language Models for Hardware Verification: A Novel SystemVerilog Assertion Dataset}


\author{Anand Menon}
\affiliation{%
  \institution{The University of Texas at Dallas}
  \city{Richardson}
  \country{USA}}

\author{Samit S Miftah}
\affiliation{%
  \institution{The University of Texas at Dallas}
  \city{Richardson}
  \country{USA}}

\author{Shamik Kundu}
\affiliation{%
 \institution{Intel Corporation}
 \country{USA}}

\author{Souvik Kundu}
\affiliation{%
 \institution{Intel Corporation}
 \country{USA}}

\author{Amisha Srivastava}
\affiliation{%
  \institution{The University of Texas at Dallas}
  \city{Richardson}
  \country{USA}}

\author{Arnab Raha}
\affiliation{%
 \institution{Intel Corporation}
 \country{USA}}

\author{Gabriel T Sonnenschein}
\affiliation{%
  \institution{The University of Texas at Dallas}
  \city{Richardson}
  \country{USA}}

\author{Suvadeep Banerjee}
\affiliation{%
 \institution{Intel Corporation}
 \country{USA}}
 
\author{Deepak Mathaikutty}
\affiliation{%
 \institution{Intel Corporation}
 \country{USA}}

\author{Kanad Basu}
\affiliation{%
  \institution{The University of Texas at Dallas}
  \city{Richardson}
  \country{USA}}
\renewcommand{\shortauthors}{Menon et al.}

\begin{abstract}

Hardware verification is crucial in modern SoC design, consuming around 70\% of development time. SystemVerilog assertions ensure correct functionality. However, existing industrial practices rely on manual efforts for assertion generation, which becomes increasingly untenable as hardware systems become complex. Recent research shows that Large Language Models (LLMs) can automate this process. However, proprietary SOTA models like GPT-4o often generate inaccurate assertions and require expensive licenses, while smaller open-source LLMs need fine-tuning to manage HDL code complexities. To address these issues, we introduce \dataset, an open-source dataset designed to enhance SystemVerilog assertion generation using LLMs. \dataset enables researchers in academia and industry to fine-tune open-source models, outperforming larger proprietary ones in both accuracy and efficiency while ensuring data privacy through local fine-tuning and eliminating costly licenses. The dataset is curated by systematically augmenting variables from open-source HDL repositories to generate synthetic code snippets paired with corresponding assertions. Experimental results demonstrate that fine-tuned models like Deepseek Coder 6.7B and Llama 3.1 8B outperform GPT-4o, achieving up to 96.88\% improvement over base models and 24.14\% over GPT-4o on platforms including OpenTitan, CVA6, OpenPiton, and Pulpissimo. \dataset is available at \href{https://github.com/AnandMenon12/VERT}{https://github.com/AnandMenon12/VERT}.
\end{abstract}






\maketitle
\input{Snippets/config}
\input{sections/introduction}
\input{sections/Background}

\input{sections/Motivation}

\input{sections/Meth}

\input{sections/Results}
\input{sections/conclusion}

\section{Acknowledgments}
This research is supported by NSF grant \#222304 and Intel Corporation. 


\printbibliography


\end{document}

%% file: Snippets/config.tex
\definecolor{codegreen}{rgb}{0,0.6,0}
\definecolor{codegreen}{rgb}{0,0.6,0}
\definecolor{codegray}{rgb}{0.5,0.5,0.5}
\definecolor{codepurple}{rgb}{0.58,0,0.82}
\definecolor{backcolour}{rgb}{1,1,1}
\definecolor{darkred}{rgb}{0.6,0,0}

\lstdefinestyle{mystyle}{
    framexleftmargin=10pt,
    frameround=tttt,
    frame=single,
    commentstyle=\color{codegreen},
    keywordstyle={\color{blue}\bfseries},
    numberstyle=\tiny\color{codegray},
    stringstyle=\color{codepurple},
    basicstyle=\ttfamily\tiny,
    breakatwhitespace=false,         
    breaklines=true,                 
    captionpos=b,                    
    keepspaces=true,                 
    numbers=left,                    
    numbersep=5pt,                  
    showspaces=false,                
    showstringspaces=false,
    showtabs=false,                  
    tabsize=1,
    xleftmargin= 15pt,
    xrightmargin= 5pt
}
 
\lstset{
style=mystyle,
emph={Module, property, endproperty, disable, iff, `PRIM_FLOP_SPARSE_FSM, unique, always\_ff, always\_comb, typedef, constraint, assert}, 
emphstyle={\color{blue}\bfseries},
emph={[2]\$stable, \$isunknown, \$isinside, \$past, \$error, \$fatal, enum, extends}, 
emphstyle={[2]\color{darkred}\bfseries}, 
emph={[3]aes\_core}, 
emphstyle={[3]\color{codepurple}\bfseries}
}

\lstdefinestyle{mystyle2}{
    backgroundcolor=\color{backcolour}, 
    framexleftmargin=10pt,
    frame=none,
    commentstyle=\color{codegreen},
    keywordstyle={\color{blue}\bfseries},
    numberstyle=\tiny\color{codegray},
    stringstyle=\color{codepurple},
    basicstyle=\ttfamily\tiny,
    breakatwhitespace=false,         
    breaklines=true,                 
    captionpos=b,                    
    keepspaces=true,                 
    numbers=left,                    
    numbersep=5pt,                  
    showspaces=false,                
    showstringspaces=false,
    showtabs=false,                  
    tabsize=1,
    xleftmargin= 15pt,
    xrightmargin= 5pt
}

%% file: sections/introduction.tex
\section{Introduction}



In modern computing, System-on-Chip (SoC) designs have become dominant, offering extensive integration of various Intellectual Property (IP) cores into a single chip \cite{miftah2024system}. While this approach significantly reduces production timelines and lowers costs, it also introduces critical challenges. One of the most pressing issues is the detection of functional bugs in these complex designs, which can consume up to 70\% of the overall development time~\cite{farahmandi2020system}. Failure to detect design bugs prior to chip fabrication can lead to significantly higher post-production costs. This emphasizes the necessity of rigorous pre-manufacturing verification processes to identify and resolve potential issues early. Early detection mitigates the need for costly redesigns and minimizes production delays. Consequently, thorough hardware verification before fabrication is essential to ensure the design operates as intended and meets performance requirements. Current industrial practices rely on the manual generation of SystemVerilog Assertions (SVA), which play a crucial role in addressing verification needs by enabling early bug detection through the capture of critical system properties. However, manually generating these assertions is a time-consuming process that depends heavily on the designer's expertise. This not only makes it challenging to adapt to complex designs but also increases the risk of human error~\cite{dessouky2019hardfails, fang2024assertllm}.

Large Language Models (LLMs) help overcome these issues by analyzing text data, like code, to automatically generate assertions. This process automates the otherwise tedious manual task of assertion writing, ensuring a significant reduction in time and human effort and leading to more efficient verification cycles. However, recent academic research shows that proprietary and open-source LLMs struggle with generating high-quality Verilog code, including assertions. Even models such as Open AI's GPT-4 perform poorly in Verilog code generation due to a lack of high-quality, model-specific tuning data \cite{zhao2024codev}. This is further substantiated by recent research, which showed that only 11\% of the assertions generated by GPT-4 on the OpenTitan SoC were unique and correct  \cite{kande2024security}.

Specifically, these LLMs often generate assertions that are neither syntactically nor functionally correct, necessitating human intervention. These issues are further elaborated in detail in Section \ref{sec:motivation}. Conversely, employing a tailored Verilog code dataset for hardware design has been shown to significantly enhance LLM generation capabilities, demonstrating the importance of curating open-source, high-quality, hardware-specific datasets to fully leverage LLMs in hardware design and verification~\cite{liu2024chipnemo}.



To this end, we introduce \dataset, a large-scale, high-quality, open-source dataset explicitly designed for formal and dynamic verification. Our dataset overcomes the high costs and restrictions of proprietary models like GPT-4o by enabling researchers and hardware companies to fine-tune smaller, open-source LLMs to outperform models like GPT-4o. The key advantage of our dataset is that it allows smaller, more efficient models to achieve higher accuracy and functionality than larger, licensed LLMs without the associated costs or restrictions. By open-sourcing \dataset, we not only enable local fine-tuning to safeguard sensitive design data but also provide a solution that enhances both performance and accessibility. \textbf{Our ultimate goal is to demonstrate that with the right dataset, even compact, open-source models can deliver superior results, offering a cost-effective and scalable foundation for automated hardware verification.}

Our work introduces several key contributions to hardware verification using LLMs:
\begin{itemize}

    \item We, for the first time, introduce \dataset, an open-source dataset specifically designed for SystemVerilog assertion generation. This dataset addresses the limitations of existing proprietary models and provides a valuable resource for advancing the hardware verification pipeline.
    



    \item \dataset was developed to address critical challenges in assertions generated by ChatGPT (GPT-4o), such as misinterpretation of clock cycle delays, incorrect mapping of `\texttt{if}' conditions to `\texttt{else}' branches, and oversimplification of long logical expressions. These limitations highlight ChatGPT's inability to accurately capture complex hardware events from RTL source code. In contrast, \dataset effectively enables open-source LLMs to mitigate these biases, making it a valuable contribution to improving assertion generation in hardware verification.


    \item \dataset comprises of various RTL code segments along with their valid formal counterparts \ie SystemVerilog assertions that ensure both syntactical and functional correctness, allowing open-source models to gather domain-specific SystemVerilog-based verification knowledge. 


    \item To demonstrate the effectiveness of \dataset, we perform a thorough verification process, including mutation testing, formal verification, and simulation-based analysis. This ensures full coverage of all generated assertions, even in asynchronous-reset scenarios, evaluating their correctness and reliability across various hardware modules.
   
    \item The development of \dataset has enabled smaller, open-source LLMs such as DeepSeek Coder 6.7B and Llama 3.1 8B to surpass much larger proprietary models like GPT-4o. These models demonstrate up to a 24.14\% improvement in generating precise assertions for industry-standard SoCs, including OpenTitan, Pulpissimo, CVA6, and OpenPiton.

\end{itemize}

The rest of this paper is structured as follows: Section~\ref{sec:preli} provides background and related work, discussing existing hardware verification approaches, LLM applications in hardware design, and datasets relevant to the field. Section~\ref{sec:motivation} highlights key challenges faced by state-of-the-art LLMs in generating SystemVerilog assertions, motivating the need for a dedicated dataset like \dataset. Section~\ref{sec:dataset} describes the \dataset dataset, detailing its construction, sources, synthetic assertion generation, and how it addresses LLM limitations. Section~\ref{sec:results} presents experimental results, evaluating fine-tuned LLMs against baseline models and GPT-4o on multiple hardware benchmarks. It also includes data leakage verification, an ablation study, and a comparative analysis of assertion correctness.

%% file: sections/Background.tex
 \section{Background and Related works}
\label{sec:preli}


\subsection{Hardware Verification}

As modern hardware designs grow in complexity, ensuring their functional correctness has become increasingly challenging~\cite{IBMcomplex}. Hardware verification plays a critical role in guaranteeing that these designs meet their specifications and remain error-free~\cite{gupta1992formal}. Two major approaches are commonly used in hardware verification: formal and simulation-based verification. Each approach leverages either a golden reference model (GRM) or assertions. GRMs are typically restricted to simulation-based verification, which simulates hardware behavior to check against expected outcomes. However, assertions offer greater flexibility, as they can be applied in formal and simulation-based verification environments~\cite{miftah2024system, zhang2018end}.

Assertions in formal verification mathematically prove whether design properties can be violated, ensuring critical behaviors are maintained. In simulation-based verification, assertions monitor execution and flag violations, helping identify errors early and reducing the risk of critical failures. Despite their importance, assertions are traditionally manually written by designers or verification engineers. This manual process is both time-consuming and prone to human error, especially in large, complex systems. The limited scalability of manually generated assertions contributes to longer development cycles and increases the risk of incomplete verification coverage, highlighting the need for automation in this domain.

\subsection{LLMs for Hardware Design}
\label{bg:LLMSHARD}

Recent studies have shown promise in employing LLMs to automate various hardware design and verification tasks. These approaches demonstrate that LLMs can significantly reduce manual effort by assisting in the generation of verification assertions, test stimuli, and security checks. However, several limitations have been identified across these works.

Recently, a framework named \textbf{AssertLLM} was developed, which explores the use of LLMs to generate and evaluate hardware verification assertions. \textbf{AssertLLM} leverages multiple LLMs to generate assertions from hardware design specifications \cite{fang2024assertllm}. However, the authors highlight that general-purpose LLMs often struggle with understanding formal verification semantics, leading to syntactically incorrect or logically incomplete assertions. Similar studies on LLM-assisted methods for automatically extracting properties from design documentation also face challenges with formal verification semantics~\cite{kande2024security}. Due to insufficient structured training on assertion-based verification, these methods often generate assertions that are either syntactically incorrect or logically incomplete. 

In the domain of hardware test stimulus generation, the framework \textbf{LLM4DV} uses LLMs to automate the generation of test cases~\cite{zhang2023llm4dv}. While this approach improves test coverage and reduces manual effort, it suffers from generating functionally invalid test stimuli because LLMs often fail to account for complex hardware interactions.  The framework \textbf{SCAR} demonstrated using LLMs for generating SystemVerilog implementations with first-order masking, a critical countermeasure against side-channel attacks~\cite{srivastava2023scar}. However, a key challenge identified in this approach was ensuring syntactic correctness and adherence to hardware design constraints. This is because LLMs often produce incomplete or structurally incorrect HDL code, which require either manual refinement or automated syntax correction mechanisms.

Similarly, LLM-driven automation to generate constraints for SystemVerilog verification workflows often misinterprets HDL syntax and semantics—leading to errors in constraint formulation—when converting natural language into HDL \cite{orenesvera2023using}. The framework \textbf{NSPG} aimed to classify properties from hardware documentation~\cite{meng2024nspg}. However, a major limitation in this process stemmed from LLMs’ lack of domain-specific understanding of SystemVerilog syntax and semantics, making direct text-to-assertion translation highly error-prone. \textbf{SoCureLLM} is another LLM-powered tool for verifying security policies within SoC architectures, although it encounters difficulties in differentiating between functional and security errors~\cite{tarek2024socurellm}.

In the context of design verification and error detection, the framework \textbf{UVLLM} integrates LLMs with the Universal Verification Methodology (UVM) to automate RTL design verification by automatically generating test benches and evaluating verification quality \cite{hu2024uvllm}. The authors reflect a shift toward the practical integration of LLMs into formal verification workflows, but they still face challenges related to model hallucinations, limited HDL syntax comprehension, and a lack of structured datasets. Finally, a comparative study assesses LLMs in assertion generation, security verification, and design validation, reinforcing that limited domain-specific training results in plausible but often incorrect assertions \cite{blocklove2024evaluating}.

In summary, while LLMs hold significant promise for automating key aspects of hardware design verification, current frameworks are hindered by issues such as (1) inadequate handling of formal verification semantics, (2) inability to model complex hardware interactions, and (3) challenges in accurately interpreting HDL constructs. A key aspect of these studies is their primary reliance on ChatGPT models, such as GPT-3.5 or GPT-4 and other subpar models such as Falcon 7B and BERT~\cite{chiang2024chatbot}. Nevertheless, even the most advanced iteration of ChatGPT, GPT-4o, encounters significant issues that can lead to syntactically or functionally incorrect outputs. These limitations are examined in further detail in Section~\ref{sec:motivation}.

\subsection{Datasets for Hardware Design}

Recent research demonstrated that many of the issues mentioned in Section~\ref{bg:LLMSHARD} can be addressed by creating datasets that provide LLMs with the necessary domain-specific knowledge \cite{liu2024chipnemo}. Building on this insight, researchers have generated several Verilog datasets to aid LLM performance in hardware design tasks.

The \textbf{MG-Verilog} dataset comprises over 11,000 Verilog code samples paired with corresponding natural language descriptions and is structured to provide varying levels of detail \cite{zhang2024mg}. The \textbf{SA-DS} dataset includes a collection of spatial array designs adhering to the standardized Berkeley’s Gemini accelerator generator template, thereby promoting LLM-driven research on deep neural network hardware accelerator architectures \cite{vungarala2024sa}. In the context of RTL design, the \textbf{RTL-Repo} benchmark evaluates LLM capabilities on large-scale RTL design projects by providing over 4,000 Verilog code samples extracted from public GitHub repositories—each annotated with full repository context to improve training and inference \cite{allam2024rtl}.

Furthermore, the framework \textbf{OpenLLM-RTL} consists of 50 hand-crafted designs with accompanying design descriptions, test cases, and correct RTL implementations. It also provides an extended dataset of 80,000 instruction-code samples (with 7,000 high-quality verified samples) to aid in LLM training and evaluation \cite{liu2024openllm}. The \textbf{VerilogEval} benchmark comprises 156 problems sourced from the HDLBits platform, covering a range of hardware design tasks—from simple combinational circuits to complex finite-state machines—and allows for automatic functional correctness testing by comparing simulation outputs of generated designs with golden solutions \cite{liu2023verilogeval}. Moreover, the framework \textbf{VeriGen}—which comprises Verilog code collected from GitHub repositories and Verilog textbooks—has been utilized to fine-tune pre-existing LLMs to generate high-quality Verilog code \cite{thakur2023verigen}.

In the domain of Electronic Design Automation (EDA), the \textbf{EDA Corpus} dataset features over 1,000 data points structured in two formats: (i) question prompts with prose answers and (ii) code prompts with corresponding OpenROAD scripts, thereby supporting LLM research in optimizing and automating EDA workflows \cite{wu2024eda}. Moreover, the \textbf{Hardware Phi-1.5B} framework contributes a tiered collection of hardware-related data—including design specifications, code samples, and documentation—to enhance LLM performance in various hardware design applications \cite{fu2024hardware}. Finally, in the realm of hardware security, the \textbf{Vul-FSM} database comprises 10,000 finite state machine (FSM) designs, each embedded with one or more of 16 distinct security weaknesses, and it was generated using the SecRT-LLM framework to efficiently insert and detect hardware vulnerabilities \cite{10545393}.

Despite the impressive progress made with these datasets and benchmarks, several challenges remain unresolved. Many of these works primarily focus on code generation, architectural optimization, or vulnerability identification rather than providing a comprehensive framework for assertion-based verification. Modern hardware designs necessitate precise and holistic verification approaches that can capture subtle timing issues, corner-case behaviors, and nuanced design intents. The existing datasets often lack the contextual depth, diverse assertion examples, and detailed annotation required to train LLMs in generating effective and reliable verification assertions. Therefore, there is a pressing need for a dedicated dataset that offers a comprehensive resource, tailored specifically for assertion-based verification. Such a dataset would bridge the gap between high-level design descriptions and the rigorous demands of formal verification, ultimately empowering LLMs to contribute more robustly to automated hardware verification workflows. To bridge this gap, \dataset is specifically designed to provide a comprehensive resource for training LLMs in generating SystemVerilog assertions, ensuring more reliable and robust hardware design automation.

%% file: sections/Motivation.tex
\section{Motivation}
\label{sec:motivation}

This section discusses challenges faced by state-of-the-art (SoTA) LLMs, like GPT-4o, in generating assertions for hardware verification. This motivates the need for a carefully curated dataset designed to improve the ability of the model to generate precise and reliable assertions.



\paragraph{Clock Cycle Misinterpretations:}

A common issue with models is their inability to capture clock-cycle delays in assignment logic accurately. LLMs often generate assertions referencing incorrect or future clock cycles, leading to faulty hardware verification. As illustrated in Textbox~\ref{tbox:sv_clkcycle}, the source code is sensitive to the clock's rising edge and the reset signal's falling edge. When a reset signal is received, assignments should occur within the same clock cycle. However, all the generated assertions use the ``$|=>$'' symbol, indicating that the assignment should be verified in the next clock cycle. This suggests a model bias towards checking assignments in the cycle following the satisfaction of the precondition. Therefore, it is necessary to fine-tune the model to enable accurate capturing of the clock cycle delay from the source code. This can be inferred from the `\texttt{always}'-block type and the sensitivity list. For example, the `\texttt{always\_comb}' block means everything under this block is combinatorial and, therefore, instantaneous. 
\input{GPT4-Listings/wrong_cond_and_clkcycle}



\paragraph{Incorrect Mapping of \textit{if} Conditions to \textit{else} Branches:}
One of the common pitfalls encountered by LLMs in assertion generation is their failure to correctly associate the logical dependencies between \textit{if} conditions and their corresponding \textit{else} or \textit{else-if} branches. When generating assertions for an \textit{else} block, the model often disregards the conditions that governed the preceding \textit{if} statements, leading to incomplete or incorrect assertions. For instance, in Textbox~\ref{tbox:sv_nestedIF}, GPT-4o generates an assertion that does not fully capture the dependencies established by the conditional logic. The second assertion should ensure that `\texttt{monitor\_enable}' is \texttt{FALSE} while `\texttt{monitor\_enable\_q}' remains \texttt{TRUE}. Additionally, when `\texttt{bus\_release\_cnt\_load}' is \texttt{TRUE}, the assertion must verify that the subsequent assignment takes place accordingly. This issue arises because LLMs tend to process each condition independently rather than treating them as part of an interdependent logical structure. As a result, they often fail to propagate constraints from prior conditions into subsequent branches. Such errors can lead to verification gaps where invalid transitions or unintended behaviors in the hardware design go undetected.

\paragraph{Nested \textit{If-Else} Conditions:}
LLMs often struggle with generating assertions for deeply nested \textit{if-else} conditions (i.e., blocks nested beyond two levels), which are common in hardware design. In Textbox~\ref{tbox:sv_nestedIF}, GPT-4o incorrectly represents the logic of a nested block. Specifically, it ignores the \textit{if} condition and creates an assertion using a ternary operator, where the inner \textit{if} condition is used as the test, the assignment as the true branch, and $1'b1$ for the false branch. This assertion has two issues: first, SVA does not support ternary operators; next, the original code lacks an \textit{else} condition, meaning the value should remain unchanged when the condition is false.

\paragraph{Handling Long Conditions:}

Another challenge with LLMs is their difficulty capturing long conditions involving multiple logical operators and variables. These complex conditions often result in overly simplistic or incorrect assertions. For instance, the assertions provided in the response in Textbox~\ref{tbox:sv_long} are incorrect, because they are structured with a logical flaw in how the conditional checks and error-handling mechanisms are designed. Specifically, the assertions only trigger an error when both the condition (\eg \texttt{key\_full\_sel\_o != mr\_key\_full\_sel[i]}) and the error flag \texttt{(mr\_err == 1'b1}) are true. This leads to an issue where the check fails to capture all error conditions properly. Ideally, the assertion should report an error when any of the conditions provided fails.


\input{GPT4-Listings/Nested_if_else}
\input{GPT4-Listings/Handling_long}

Therefore, addressing these challenges through a structured dataset is essential to improving the LLM's ability to generate accurate SystemVerilog assertions, ensuring better alignment with the underlying hardware design logic. Illustration of \dataset to addressing these challenges are presented in Section \ref{vertvsgpt4}.

%% file: GPT4-Listings/wrong_cond_and_clkcycle.tex
\begin{mybox}[colback=white, colframe=black, boxrule=0.2mm, sharp corners, coltitle=black, colbacktitle=white, label=tbox:sv_clkcycle]{Clock Cycle and Pre-condition Capturing in GPT-4o Outputs.}
Generate a list of synchronous SystemVerilog assertions executing at \texttt{(\textcolor{blue}{\textit{posedge}} clk\_i)} from the following code:
\lstset{style=mystyle2}
\begin{lstlisting}[language=Verilog][h]
always @(posedge clk_i or negedge rst_ni) begin
    if (~rst_ni) begin
        wr_state_q <= IDLE;
        wr_cnt_q   <= '0;
    end else begin
        wr_state_q <= wr_state_d;
        wr_cnt_q   <= wr_cnt_d;
    end
end
\end{lstlisting}

Response:

\begin{lstlisting}[language=Verilog]
...
// Assertion to ensure that wr_state_q follows wr_state_d when rst_ni is high
assert property (@(posedge clk_i) (rst_ni && (wr_state_d !== IDLE)) |=> (wr_state_q == wr_state_d))
  else $error("State transition failure: wr_state_q did not follow wr_state_d.");

// Assertion to ensure that wr_cnt_q follows wr_cnt_d when rst_ni is high
assert property (@(posedge clk_i) (rst_ni && (wr_cnt_d !== '0)) |=> (wr_cnt_q == wr_cnt_d))
  else $error("Counter transition failure: wr_cnt_q did not follow wr_cnt_d.");

// Assertion to check that the state and count do not update during reset
assert property (@(posedge clk_i) (~rst_ni) |=> (wr_state_q == IDLE && wr_cnt_q == '0))
  else $error("Reset condition not maintained: wr_state_q or wr_cnt_q changed during reset.");
...
\end{lstlisting}

\end{mybox}


%% file: GPT4-Listings/Nested_if_else.tex
\begin{mybox}[colback=white, colframe=black, boxrule=0.2mm, sharp corners, coltitle=black, colbacktitle=white, label=tbox:sv_nestedIF] {Relation Representation for Nested Conditional Blocks in GPT-4o Outputs.}
Generate a list of asynchronous SystemVerilog Assertions from the following code : 
\lstset{style=mystyle2}
\begin{lstlisting}[language=Verilog]
if (!rst_ni) begin
      bus_release_cnt <= '0;   
end else if (monitor_enable &&!monitor_enable_q) begin
    if (multi_controller_enable_i) begin
        bus_release_cnt <= 30'(bus_inactive_timeout_i);
    end
end else if (bus_release_cnt_load) begin
    bus_release_cnt <= bus_release_cnt_sel;
end else if (bus_release_cnt_dec && (bus_release_cnt!= '0)) begin
    bus_release_cnt <= bus_release_cnt - 1'b1;
end
\end{lstlisting}

Response:

\begin{lstlisting}[language=Verilog]
...
// Synchronous assertion for monitor enable signal transition
assert property (@(posedge clk) disable iff (!rst_ni)
    (monitor_enable && !monitor_enable_q) |-> (multi_controller_enable_i ? 
    (bus_release_cnt == bus_inactive_timeout_i) : 1'b1)
) else $fatal("Bus release count should be set to bus_inactive_timeout_i when monitor_enable rises and multi_controller_enable_i is high.");

// Synchronous assertion for bus_release_cnt_load condition
assert property (@(posedge clk) disable iff (!rst_ni)
    (bus_release_cnt_load) |-> (bus_release_cnt == bus_release_cnt_sel)
) else $fatal("Bus release count should be loaded with bus_release_cnt_sel when bus_release_cnt_load is asserted.");
...
\end{lstlisting}

\end{mybox}




%% file: GPT4-Listings/Handling_long.tex
\begin{mybox}[colback=white, colframe=black, boxrule=0.2mm, sharp corners, coltitle=black, colbacktitle=white, label=tbox:sv_long]{Handling Long Conditions in GPT-4o Outputs.}
Generate a list of asynchronous SystemVerilog Assertions from the following code:
\lstset{style=mystyle2}
\begin{lstlisting}[language=Verilog]
     if (state_sel_o      != mr_state_sel[i]     ||
          add_rk_sel_o    != mr_add_rk_sel[i]    ||
          key_full_sel_o  != mr_key_full_sel[i]  ||
          key_dec_sel_o   != mr_key_dec_sel[i]   ||
          key_words_sel_o != mr_key_words_sel[i] ||
          round_key_sel_o != mr_round_key_sel[i]) begin
        mr_err = 1'b1;
      end
\end{lstlisting}

Response:

\begin{lstlisting}[language=Verilog]
...
// Asynchronous assertion to check key_full_sel_o against mr_key_full_sel[i]
assert property ((key_full_sel_o != mr_key_full_sel[i]) |-> (mr_err == 1'b1))
  else $error("Asynchronous check failure: key_full_sel_o does not match mr_key_full_sel[%0d] and mr_err is not asserted.", i);

// Asynchronous assertion to check key_dec_sel_o against mr_key_dec_sel[i]
assert property ((key_dec_sel_o != mr_key_dec_sel[i]) |-> (mr_err == 1'b1))
  else $error("Asynchronous check failure: key_dec_sel_o does not match mr_key_dec_sel[%0d] and mr_err is not asserted.", i);

// Asynchronous assertion to check key_words_sel_o against mr_key_words_sel[i]
assert property ((key_words_sel_o != mr_key_words_sel[i]) |-> (mr_err == 1'b1))
  else $error("Asynchronous check failure: key_words_sel_o does not match mr_key_words_sel[%0d] and mr_err is not asserted.", i);
...

\end{lstlisting}

\end{mybox}

%% file: sections/Meth.tex
\section{Proposed \dataset Dataset}
\label{sec:dataset}




Our proposed dataset \dataset, is a curated collection of Verilog/SystemVerilog code snippets paired with SystemVerilog assertions, designed to fine-tune LLMs for generating syntactically and functionally correct assertions for hardware verification. By addressing the biases and errors with existing LLM-generated assertions (as mentioned in Section ~\ref{sec:motivation}), we aim to improve the LLM's handling of complex SystemVerilog assertions and enhance the overall reliability of its outputs.

The rest of this section is structured as follows: Section \ref{intuit} introduces the intuition behind dataset formulation, explaining how \dataset addresses common LLM errors, such as clock cycle misinterpretations, incorrect conditional mapping, nested if-else handling, and long condition processing. Section \ref{sec:DatasetComposition} discusses the composition of \dataset. It also describes data sources, including open-source SoC repositories and synthetically generated variables, to enhance generalization. Section \ref{sec:syn_meth} explains the synthetic generation of assertions, detailing how assertion structures are systematically created from conditional logic.

\subsection{Intuition in Dataset Formulation}
\label{intuit}


\paragraph{Clock Cycle Misinterpretations:} To resolve clock cycle misinterpretation, we standardized our format by using the overlapping implication symbol ($|->$) with a specified delay count, replacing the non-overlapping symbol ($|=>$). This approach directs the LLMs' focus solely on identifying delays, thereby simplifying their task. Moreover, \dataset includes delayed assertion checks, facilitating the accurate extraction of clock cycle information from the source code.


\paragraph{Incorrect Mapping of \textit{if} Conditions to \textit{else} Branches:}

\dataset addresses the common omission of conditions in the \textit{else/else-if} branches of \textit{if-else} statements by exposing the model to diverse conditional structures, ensuring it accurately captures prior conditions when generating assertions. By incorporating examples where each \textit{else} or \textit{else-if} branch accounts for all preceding \textit{if} conditions, the dataset trains LLMs to recognize the logical flow between branches. This enhances the model’s ability to maintain logical consistency, leading to more accurate and complete assertion generation for conditional logic.

\paragraph{Nested \textit{If-Else} Conditions:} 

To address the challenge of LLMs struggling with deeply nested \textit{if-else} statements, we expanded our dataset to include complex, multi-level conditional structures. These examples focused specifically on scenarios where decision logic is nested beyond two levels, which is common in hardware designs but difficult for LLMs to handle. By providing a diverse set of deeply nested \textit{if-else} conditions, we aim to enhance the LLM's ability to better recognize how each layer of decision-making is dependent on the preceding conditions. This approach ensures that the LLM generates assertions for each nested block without oversimplifying the logic or missing critical conditions in the inner branches. \textbf{Furthermore, we refined the dataset to ensure that the LLM learns to correctly generate assertions even when the code lacks an explicit \textit{else} branch, preserving the intended behavior of the original code.} This ensures that if the condition is false, no action is required, and the state remains unchanged. However, LLMs can struggle with this distinction, often generating incorrect assertions by either assuming an implicit \textit{else} branch or failing to account for the absence of any action when the condition evaluates to false. This process helps the model handle nested structures more effectively, producing accurate and logically consistent assertions for even the most complex hardware designs.

\paragraph{Handling Long Conditions:} To address the challenge of generating accurate assertions for long and complex conditions, we expanded the dataset to include a variety of cases where multiple conditions and operators must be evaluated simultaneously. These conditions often involve a combination of AND, OR, and NOT operators across several variables, making it essential for the model to handle intricate logical relationships. By exposing the LLM to examples that require the correct ordering and evaluation of these operators, \dataset helps it learn to generate assertions that accurately reflect the complexity of the source code. \textit{This approach ensures that all logical paths are captured in the assertions, avoiding the common pitfall of oversimplifying or omitting important parts of the condition}. The result is more precise handling of extended logic chains, leading to fewer errors in assertion generation for complex hardware designs.

\begin{figure}[b]
    \centering
    \begin{subfigure}[c]{0.39\linewidth}
            \includegraphics[width=\textwidth]{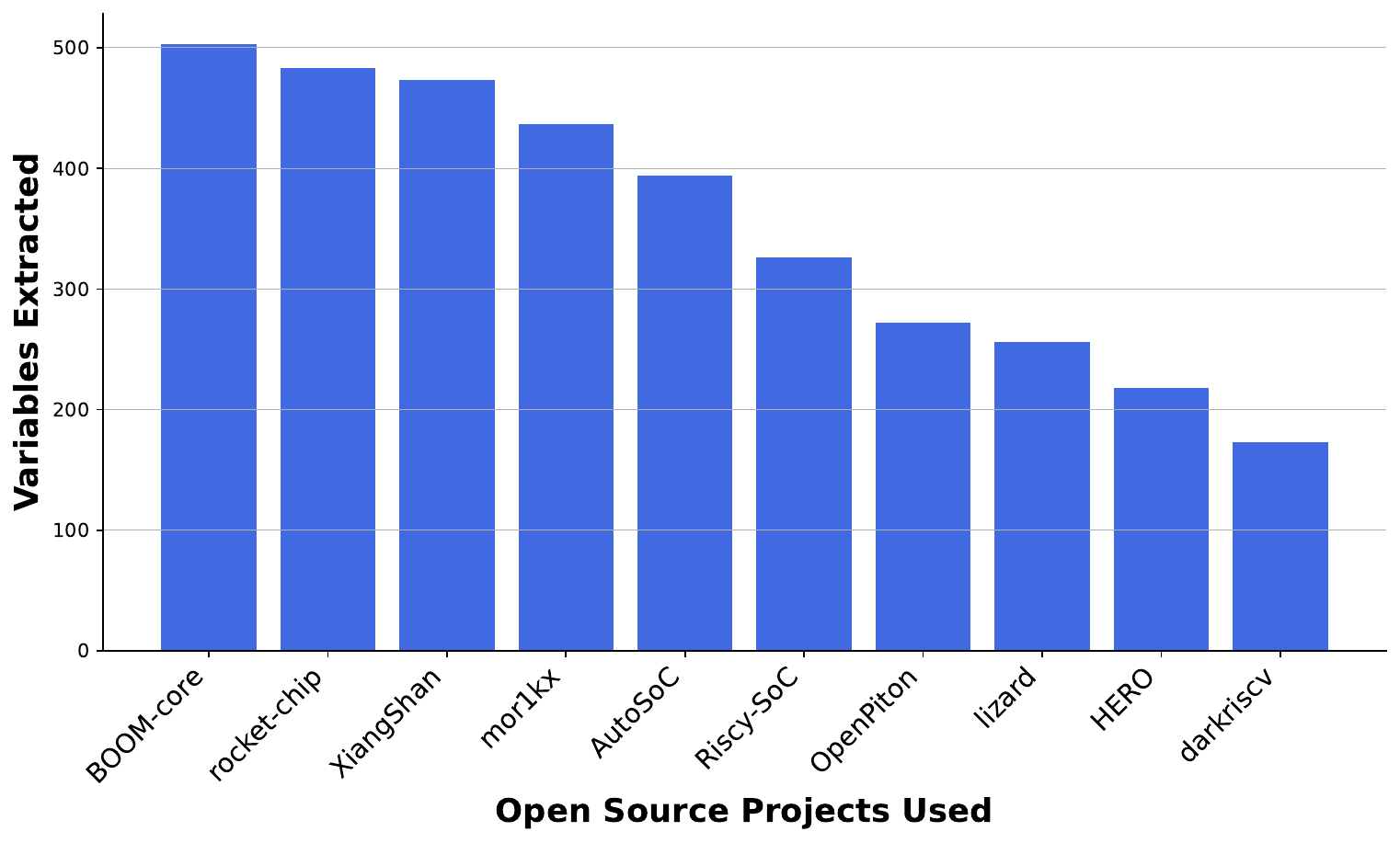}
            \caption{Distribution of variables extracted from open-source HDL projects.}
            \label{fig:OSV}
    \end{subfigure}
    \hfill
    \begin{subfigure}[c]{0.6\linewidth}
        \includegraphics[width=\linewidth] {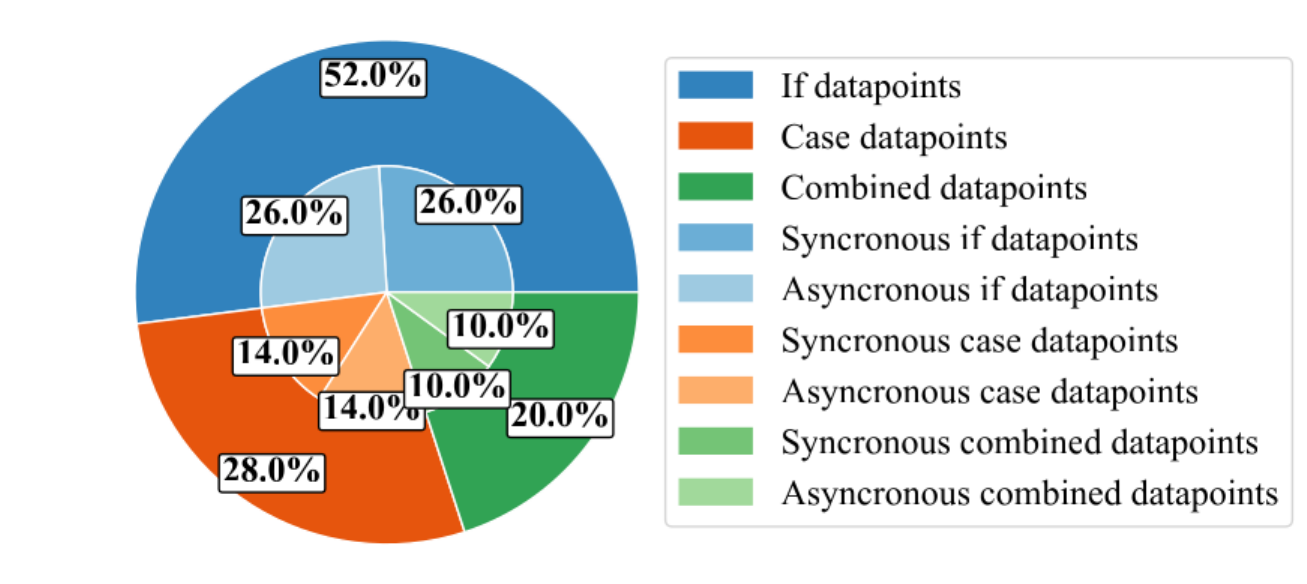}
        \caption{Composition of our dataset.}
        \label{fig:comp}
    \end{subfigure}
    \caption{Dataset source distribution and composition.}
    \label{fig:dataComp}
\end{figure}

\subsection{Dataset Composition}
\label{sec:DatasetComposition}

\dataset comprises 20,000 samples, categorized based on the structural elements of SystemVerilog code and the nature of the assertions generated. We carefully divide \dataset among various categories to ensure comprehensive coverage of the conditions encountered in hardware verification while addressing the weaknesses of current SoTA LLMs in generating assertions. Figure~\ref{fig:dataComp} illustrates the dataset's source distribution and composition. Figure 1(a) presents the distribution of variables extracted from various open-source HDL projects, showcasing the diversity of sources used in constructing the dataset. Figure 1(b) provides a breakdown of the dataset composition, categorizing data points into \textit{if-else} structures, Case structures, and Combined structures. Additionally, the dataset is further subdivided into synchronous and asynchronous variants.

\paragraph{Data Source and Cleanup:}

We compile a comprehensive list of variable names for \dataset by extracting variables from hardware modules in various open-source Hardware Description Language (HDL) projects. As shown in Figure \ref{fig:OSV}, these variables are sourced from a diverse set of RISCV projects, including \textit{BOOM-core}~\cite{zhaosonicboom}, \textit{rocket-chip}~\cite{lee2016agile}, and \textit{XiangShan}~\cite{micro2022xiangshan}, each contributing over 150 variables to the dataset. \textit{BOOM-core} leads with approximately 500 variables, while \textit{rocket-chip} and \textit{XiangShan} contribute around 450 variables each. By drawing from a diverse range of open-source modules, we ensure the model is exposed to various real-world scenarios. Many System-on-Chip (SoC) designs frequently reuse IP blocks from the same vendors, resulting in overlapping variable names. Similar IP blocks, such as various implementations of AES encryption, often perform identical operations, further contributing to naming redundancies. This reuse of IP, prevalent in both open-source and commercial SoCs, creates a degree of homogeneity in the design landscape, making it challenging to differentiate between components. To mitigate this issue and prevent overfitting to specific naming conventions or operations, we introduce randomly generated variables into the dataset, ensuring greater diversity and robustness in handling various designs. These randomly generated variables were created by algorithmically combining common hardware-related prefixes (\eg ``reg'', ``ctrl'', or ``temp'') with randomly generated alphanumeric suffixes. Once the variable list is compiled, it is cleaned up by removing duplicates, resolving inconsistencies, and verifying syntactic correctness. This ensures the model is exposed to various real-world hardware design scenarios while avoiding overfitting.

\paragraph{Rationale Behind Data Composition:}

As illustrated in Figure \ref{fig:comp}, the largest portion of the dataset, comprising 52\%, consists of \textit{if-else} statements. This focus stems from the challenges LLMs like GPT-4o often face in generating accurate assertions for nested \textit{if} structures, as discussed in Section \ref{sec:motivation}. Building upon the intuition presented in Section \ref{intuit}, we structured the dataset to prioritize complex conditional scenarios. The complexity and layering of conditions in nested \textit{if} statements frequently lead to errors, making them more problematic than other conditional structures. To address these issues, we emphasize \textit{if-else} statements in our dataset. In contrast, case statements make up 28\% of the dataset. Although commonly used to represent signals in hardware design, we encounter fewer difficulties when generating assertions for case-based logic, which accounts for their smaller proportion. Furthermore, 20\% of the dataset includes combined statements, where \textit{if} and \textit{case} statements are intertwined to form more complex conditions. These mixed scenarios are included due to the added complexity, which presents challenges for LLMs when generating accurate assertions. \textbf{We also include an even distribution of asynchronous and synchronous assertions in our dataset.} This is crucial because for LLMs to capture clock cycle delays accurately, they must correctly interpret which signals in \textit{if-else} or \textit{case} statements are clock-sensitive. By providing a mix of both types of assertions, we ensure that the models learn to differentiate between immediate and clocked responses, enabling accurate assertion generation in clock-sensitive hardware designs.




\paragraph{Structural Components in Assertion Formulation:}


To build upon the analysis from Figure~\ref{fig:comp}, where we emphasized the inclusion of various conditional structures in the dataset, it is important to highlight how these structures are integral to formulating assertions. 
When an assertion is formulated, the conditional structure of the function is required. These structures are constructed using \textit{if-else} blocks, \textit{case}, and \textit{ternary} operators. The sensitivity (\ie when to check values for assertions) is taken from the \textit{always} block. For instance, \texttt{always @(posedge clk\_i)} denotes that the values should be checked at the rising edge of the \texttt{clk\_i} signal. Our dataset contains all types of \textit{always} blocks used in hardware design codes (\ie \texttt{always}, \texttt{always\_ff}, \texttt{always\_comb}). Other code components like \textit{for} loops do not contribute to the formulation of assertions. This is because \textit{for} loops are designed for handling repetitive operations, whereas assertions are specifically intended to verify and manage the dynamic control flow of hardware.

\subsection{Synthetic Generation of Assertions}
\label{sec:syn_meth}

The proposed \dataset dataset was synthetically generated to address the variability in how different repositories and projects formulate assertions. Many open-source repositories employ custom or project-specific assertion structures, leading to inconsistencies across sources. This lack of standardization makes it challenging to compile a cohesive dataset using only real-world examples. Moreover, relying solely on real-world data would not provide a sufficient number of consistent assertion structures for an LLM to effectively learn how to generate assertions from source code. Therefore, synthetic data is essential to create a comprehensive and uniform dataset suitable for training.

Generating the synthetic data involves creating a comprehensive set of conditions based on the cleaned variable list. These conditions serve as the foundation for creating structured code blocks, along with their corresponding assertions. By dynamically generating these conditions, we can ensure that the model is exposed to a wide array of patterns, preventing it from overly relying on specific naming conventions or design features commonly encountered in available open-source SoC components.

\begin{figure}[h]
\centering
\includegraphics[width = \linewidth] {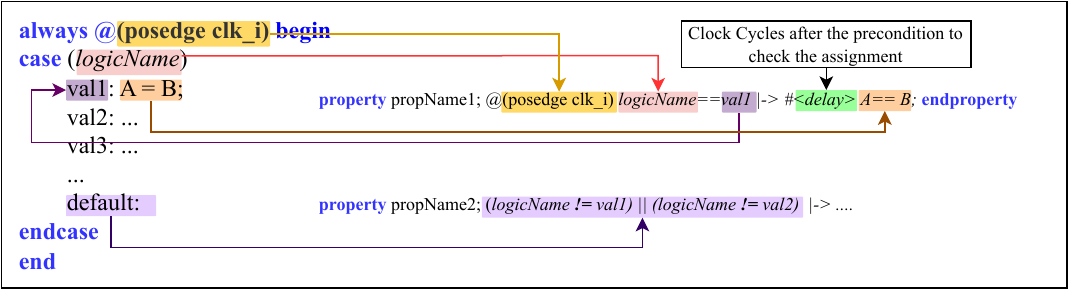}
\caption{Generation of assertions from Case Statements.}
\vspace{-2mm}
  \label{flo:case}
\end{figure}

Figure~\ref{flo:case} showcases the generation of synthetic \textit{case} statements and their corresponding assertions. The process operates by extracting select lines from a dataset of variables and conditions, and for each line, it constructs a Verilog-style \textit{case} block. It selects unique conditions and populates assignment operations. Since the conditional statements and assignment operation are known during dataset generation, the assertions can be constructed based on these conditions, ensuring consistency. In this process, the assertions are triggered on the rising edge of the clock (as indicated by the \texttt{@posedge clk\_i} in the source code), ensuring that the logic is evaluated synchronously. The selected \textit{case} checks the assigned condition, while subsequent cases ensure the appropriate actions for other input values. The default clause handles situations where none of the specified cases are met. Each \textit{case} condition is followed by a delay to account for signal propagation and verify that the expected logic occurs at the correct time. Assertions for unselected cases confirm that invalid branches are not mistakenly triggered, ensuring the default behavior is correctly executed.

\begin{figure}[h]
\centering
\includegraphics[width = \linewidth] {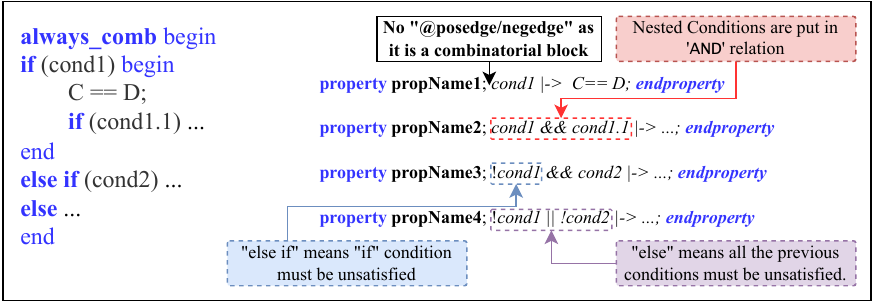}
\caption{Generation of assertions from If statements.}
\vspace{-4mm}
  \label{flo:if-else}
\end{figure}

Figure~\ref{flo:if-else} demonstrates how a hierarchy of synthetic asynchronous \textit{if-else} conditions are systematically transformed into assertions that verify the correctness of combinational logic. Since there is a combinatorial block (\texttt{always\_comb}) in the source code, the assertion created is asynchronous, hence devoid of a clock signal. The initial condition checks the first case, while nested conditions introduce additional layers of complexity. The \textit{else-if} and \textit{else} clauses account for alternative scenarios when the previous conditions are unsatisfied. In this logical flow, nested conditions are connected using an \texttt{AND} relation, requiring all specified conditions to be true for their corresponding assertions to activate. For the \textit{else-if} and \textit{else} branches, previous conditions are negated, ensuring the new condition only holds when prior conditions are false. This comprehensive approach effectively tests both \textit{if} and \textit{else} branches within the \textit{if-else} block, providing thorough coverage of all possible logical states.

%% file: sections/Results.tex
\section{Results}
\label{sec:results}

In this section, we present a comprehensive evaluation of our dataset, \dataset, and its impact on hardware verification through a series of methodical experiments. 

We begin in Section~\ref{sec:expsetup} by detailing our experimental setup, which outlines the hardware platform, the selected LLMs to fine-tune using \dataset, and the open-source SoC benchmarks used in our evaluations. Section~\ref{sec:metrics} introduces our evaluation metrics, focusing on the methods used for checking syntactical and functional correctness. To ensure the integrity of our evaluation framework, Section~\ref{sec:leakage} addresses potential data leakage concerns. Section~\ref{sec:fine_tune} then outlines the fine-tuning hyperparameters employed—detailing our use of the Low-Rank Adapter (LoRA) technique.

The core evaluation results are presented in Section~\ref{Sec:EvalResults}. This section compares the performance of models before and after being trained on \dataset in generating syntactically and functionally correct assertions, benchmarks these results against GPT-4o, and includes a coverage measurement discussion that validates the thoroughness of the assertion generation by the LLMs fine-tuned on \dataset. An ablation study is detailed in Section~\ref{sec:abl_study}, where we analyze the impact of uncleaned versus cleaned variable names on assertion quality. Section~\ref{sec:label} provides qualitative examples that illustrate how LLMs fine-tuned on VERT overcome common pitfalls observed in GPT-4o and also offers a comparative analysis with human expert assertions—highlighting improvements in handling clock cycles, nested conditions, and long, complex conditions. Finally, Section~\ref{sec:cont} investigates the impact of contamination in design files on the accuracy of assertion generation, providing insights into the robustness of our approach under challenging conditions.



\subsection{Experimental Setup}
\label{sec:expsetup}

In our experiments, we employed two prominent open-source large language models—LLama 3.1 8B and DeepSeek Coder 6.7B—to assess code generation for hardware design verification tasks \cite{dubey2024llama, guo2024deepseek}. LLama 3.1 was chosen for its efficiency and robust performance across various natural language processing tasks, offering rapid inference and resource efficiency that facilitate swift experimental iterations. In contrast, DeepSeek Coder has been specifically fine-tuned for code generation, demonstrating a strong aptitude for producing syntactically correct and functionally relevant code. By leveraging these two models, we aim to benchmark the effectiveness of our dataset, \dataset, across both general-purpose and code-specialized architectures, thereby providing a balanced assessment of diverse fine-tuning strategies.

Furthermore, since GPT-4o cannot be fine-tuned as it is a closed-sourced model, we compare the open-source fine-tuned models to GPT-4o to highlight the effectiveness of \dataset. In order to assess their ability to generate code across diverse coding conventions and design principles, we test them on three open-source SoC designs—OpenTitan \cite{opentitan_2024}, CVA6 \cite{zaruba2019cost}, Pulpissimo \cite{8640145}, and OpenPiton \cite{balkind2016openpiton}. Our experimental platform comprises an Nvidia DGX server equipped with four Nvidia A-100 80GB GPUs.

\subsection{Evaluation Metrics} 
\label{sec:metrics}
The evaluation focuses on three primary metrics: (1) the total number of generated assertions, (2) the percentage of syntactically correct assertions, (3) the percentage of functionally correct assertions, and (4) the Complete Path Coverage (CPC) of the assertions. Syntactic correctness refers to adherence to hardware description language standards, while functional correctness indicates that the assertions accurately reflect intended hardware behavior. CPC refers to covering all possible independent paths within an automaton. A path begins at the initial node, traverses through the graph's edges, and ends at a final node.


Textbox~\ref{tb:diff_func_syn} further clarifies the difference between syntactical and functional assertions by illustrating the distinct errors each type can incur within a hardware design context. Specifically, it outlines two common issues that may arise when writing assertions to verify system behavior. 

\input{Appendix_syn_func/textbox}

\paragraph{Design Source Code:}
The provided SystemVerilog code in Textbox \ref{tb:diff_func_syn} shows an \texttt{always\_ff} block, triggered by either the rising edge of \texttt{clk\_aon\_i} or the falling edge of \texttt{rst\_aon\_ni}. The block resets or updates the value of \texttt{wakeup\_timer\_cnt\_q} based on certain conditions:
\begin{enumerate}
    \item If \texttt{rst\_aon\_ni} is low (reset active), the counter is set to zero.
    \item If \texttt{wakeup\_timer\_cnt\_clr} or certain other signals are asserted, the counter is reset.
    \item Otherwise, the counter is updated with a new value from \texttt{wakeup\_timer\_cnt\_d}.
\end{enumerate}

\paragraph{Assertions:}
Two properties are presented, each demonstrating a different type of error:

\begin{itemize}
    \item \textbf{Syntactical Error:} \\
    In \texttt{property ResetTimer1}, the assertion attempts to check if the counter is reset when \texttt{rst\_aon\_ni} is low. However, it contains a \textbf{syntactical error}: instead of using the comparison operator \texttt{==} to check if \texttt{wakeup\_timer\_cnt\_q} equals zero, the assignment operator \texttt{=} is mistakenly used. This would result in a syntax error during compilation.
    
    \item \textbf{Functional Error:} \\
    In \texttt{property ResetTimer2}, while the syntax is correct, the assertion misses an essential condition. It checks whether the counter is reset when the clear signal or related signals are asserted. However, it fails to include the reset condition (\textit{i.e.,} \texttt{if (!rst\_aon\_ni)}), resulting in a \textbf{functional error} because the assertion does not fully capture the intended behavior of the design, specifically missing the \texttt{if} condition from the original source code.
\end{itemize}

\input{Appendix_syn_func/keypoint}

We evaluate the functional correctness of the generated assertions through a two-step process that combines mutation testing with formal and simulation-based verification. First, we introduce mutation testing—consistent with the methodology in \cite{iman2024artmine}—by intentionally inserting small, targeted code modifications (mutants) into the design. These mutants deviate from the expected assertion logic, serving as benchmarks to verify that the assertions are effective in detecting logical inconsistencies. If a mutant does not trigger the corresponding assertion, that assertion is deemed functionally incorrect.

Next, we employ Cadence JasperGold, a state-of-the-art commercial formal verification tool, to check whether the mutations trigger the assertions~\cite{jasper}. JasperGold’s speed and formal analysis capabilities allow us to efficiently identify any functional errors. However, formal tools like JasperGold assume that asynchronous reset signals remain inactive during execution \cite{miftah2024system}, which means they cannot verify properties that involve asynchronous resets.

 To address this limitation, we simulate the mutated design using Xilinx Vivado~\cite{vivado}. This simulation step ensures that assertions involving asynchronous resets are adequately covered, thereby complementing the formal verification stage. By combining mutation testing with both formal verification via JasperGold and simulation via Vivado, we ensure that only syntactically correct and functionally valid assertions are retained in the design.

This comprehensive approach not only confirms the robustness and relevance of the generated assertions—demonstrated by their high mutation detection rates across various benchmarks—but also guarantees complete coverage of conditional branches and critical logic paths within the hardware design.

\subsection{Data Leakage Verification}
\label{sec:leakage}
\input{Tables/LeakageTable}


Data leakage between the evaluation framework and \dataset is a critical concern as it can lead to overestimated performance and misleading evaluations of model generalization. Preventing leakage is essential to ensure that the experimental results faithfully represent real-world performance and that benchmark comparisons remain fair. To ensure that no inadvertent data leakage occurs between \dataset and the evaluation benchmarks, we adopt a data leakage detection method based on byte-level 13-gram overlap~\cite{brown2020language}. A 13-gram is a contiguous sequence of 13 bytes extracted from the text. By sliding a fixed-size window over the raw data, we capture overlapping segments that serve as unique fingerprints for comparing content across documents. This approach, similar to that employed during the development of GPT-3, operates directly on the raw byte representations of the text, thus avoiding potential inconsistencies introduced by tokenization or encoding differences.

To further elaborate, we represent a given text as an ordered sequence of bytes, as formalized in Equation~\ref{eqn:byteSeq}.

\begin{equation}
    b = (b_1, b_2, \ldots, b_L)
    \label{eqn:byteSeq}
\end{equation}

Here, \(L\) denotes the length of the byte sequence. From this sequence, we extract all contiguous 13-grams, which are defined as in Equation~\ref{eqn:13gramsSeq}.

\begin{equation}
    N(b) = \{ (b_i, b_{i+1}, \ldots, b_{i+12}) \mid 1 \leq i \leq L-12 \}.
    \label{eqn:13gramsSeq}
\end{equation}

Let \(N_1\) and \(N_2\) represent the sets of 13-grams derived from two distinct corpora (for example, our dataset and a benchmark). To quantify the degree of overlap between these sets, we employ the Jaccard similarity coefficient~\cite{jaccard1908nouvelles}, which is defined in Equation~\ref{eqn:JaccardSimilarity}:

\begin{equation}
    \text{Overlap}(N_1, N_2) = \frac{|N_1 \cap N_2|}{|N_1 \cup N_2|}
    \label{eqn:JaccardSimilarity}
\end{equation}

In this formulation, \(|N_1 \cap N_2|\) denotes the number of 13-grams common to both datasets, while \(|N_1 \cup N_2|\) represents the total number of unique 13-grams present in either dataset.

A high overlap score (\ie a value closer to 1) indicates a significant reuse of byte-level sequences between the datasets, which may be suggestive of potential data leakage. Conversely, a low overlap score (\ie a value closer to 0) implies that the datasets are largely distinct, thereby mitigating concerns regarding leakage.

The results of our data leakage evaluation are summarized in Table~\ref{tab:LeakageTable}. The table is structured with the first column listing the benchmark or hardware IP under investigation. The subsequent four columns report the computed overlap scores using the byte-level 13-gram approach applied to different components: the \textit{Code Overlap Score} quantifies the overlap in the code segments; the \textit{Assertions Overlap Score} captures the overlap in assertion texts; and the \textit{DeepSeek Coder Prompt Overlap Score} as well as the \textit{LLama 3.1 Prompt Overlap Score} reflect the overlap in the respective prompt texts generated by these systems.

Examining the results, we observe that all overlap scores are very low, typically on the order of \(10^{-4}\). For instance, the \emph{Code Overlap Scores} across various benchmarks range from 0.0001 to 0.0005, which suggests minimal commonality in the underlying code. The \emph{Assertions Overlap Score} is consistently 0 for all entries, indicating no detectable overlap in the assertion-related texts. Similarly, the overlap scores for both the DeepSeek Coder and LLama 3.1 prompts remain comparably low, with only minor variations observed across different benchmarks. These uniformly low scores confirm that there is negligible reuse of content between our dataset (\dataset) and the benchmark sources, reinforcing the integrity of our evaluation framework against data leakage.
\input{Appendix_syn_func/keypoint_overlap}

\input{Tables/hyperparam}
\subsection{Fine-Tuning Hyperparameters}
\label{sec:fine_tune}


The hyperparameters we used are listed in Table~\ref{tab:param}. Furthermore, we applied the Low-Rank Adapter (LoRA) technique with a rank and alpha of 256. This approach optimized the model for hardware verification tasks, enabling efficient low-rank updates while keeping computational overhead minimal. A maximum sequence length of 4096 tokens was employed to accommodate longer logic and condition sequences in assertion generation, with training constrained to 3 epochs to avoid overfitting.  We selected a batch size of 64 and a learning rate of $1.00e^{-04}$ for computational efficiency and stable convergence, with Bf16 (1 sign bit, 8 bits for the exponent, and 7 bits for the mantissa) precision enhancing training speed without compromising accuracy. By targeting the \texttt{Query}, \texttt{Key}, \texttt{Value}, \texttt{Output}, and \texttt{Gate} layers, we effectively adapted the model while updating only 4-6\% of its parameters, thus optimizing performance while controlling computational costs.

\subsection{Evaluation Results}
\label{Sec:EvalResults}

\subsubsection{Syntactical and functional correctness of generated assertions:}

\input{Tables/asserttable}

Table \ref{tab:asserttable} compares the performance of the base Llama 3.1 and DeepSeek Coder models with their counterparts fine-tuned on \dataset across various hardware IP benchmarks. In this context, \textit{base} refers to the original pre-trained models that were trained on broad, general-purpose datasets. In contrast, \textit{fine-tuned} models have undergone additional training on the proposed specialized dataset, \dataset. The first column of the table lists the benchmark name, such as OpenTitan/AES, where the SoC name (OpenTitan) is followed by the specific IP name (AES). The subsequent columns display the number of assertions generated and the percentage of those assertions that are both syntactically and functionally correct for both base and fine-tuned models. The table is organized by model type, with performance metrics broken down for each hardware IP block. Both the Llama 3.1 and DeepSeek Coder models demonstrated significant improvements over the base models following fine-tuning, with some benchmarks showing drastic gains. For Llama 3.1, syntactic correctness saw a maximum improvement of up to 83.33\%. Similarly, the functional correctness showed a maximum increase of 93.55\%. The DeepSeek Coder model exhibited similarly substantial improvements. For instance, syntactic correctness improved as much as 93.75\% (from 6.25\% to 100\%), and functional correctness increased up to 96.88\% (from 0\% to 96.88\%). These results highlight the effectiveness of fine-tuning in improving the models' ability to generate accurate hardware assertions.

\subsubsection{Comparison of LLMs fine-tuned on \dataset with GPT-4o:}
\begin{figure}[h]
    \begin{subfigure}[c]{0.49\textwidth}
        \includegraphics[width = \linewidth] {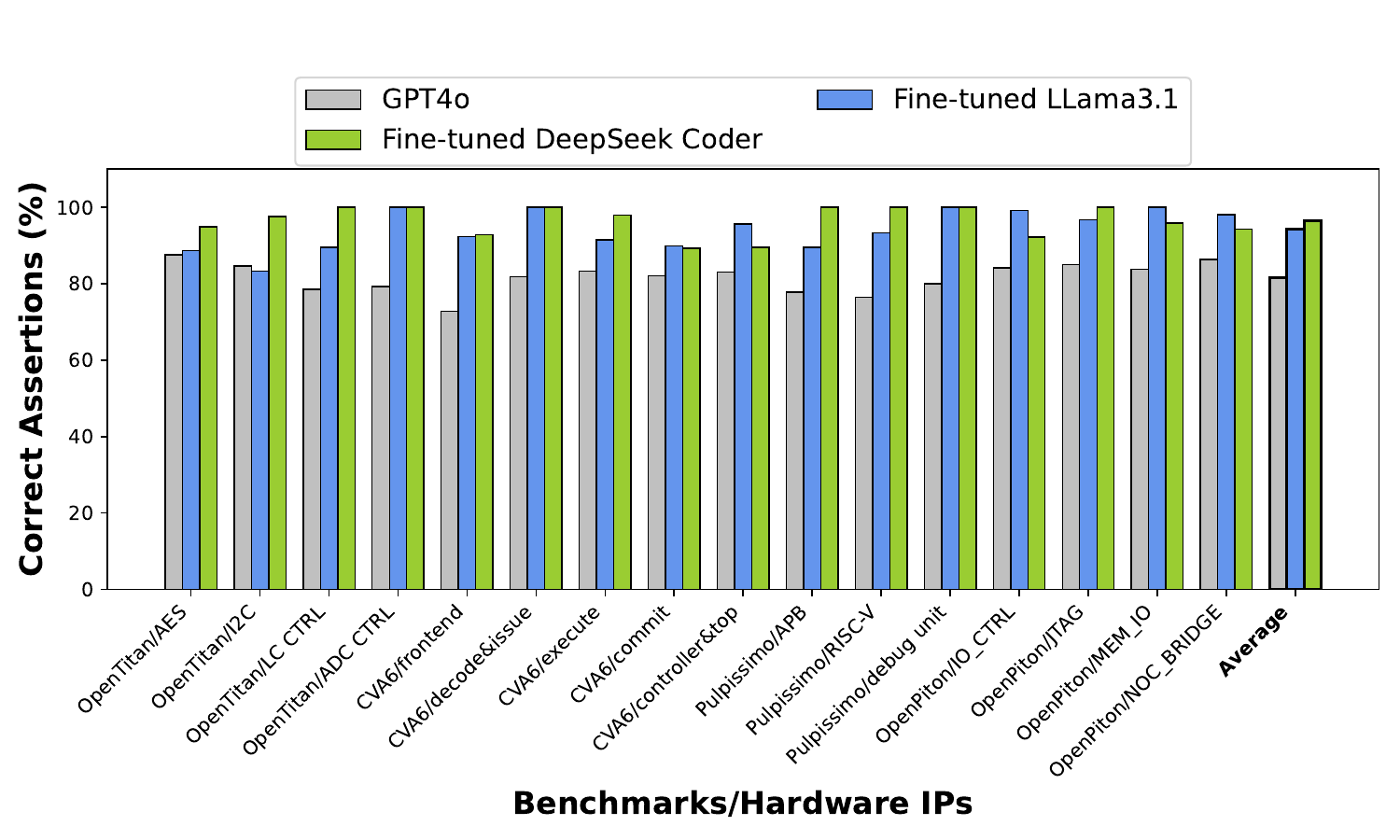}
        \caption{Comparison of GPT-4o and Fine-tuned model in Syntactically correct assertions.}
        \label{fig:gpt4_comp_syn}
    \end{subfigure}
    \hspace{1mm}
    \begin{subfigure}[c]{0.49\textwidth}
        \includegraphics[width=\textwidth]{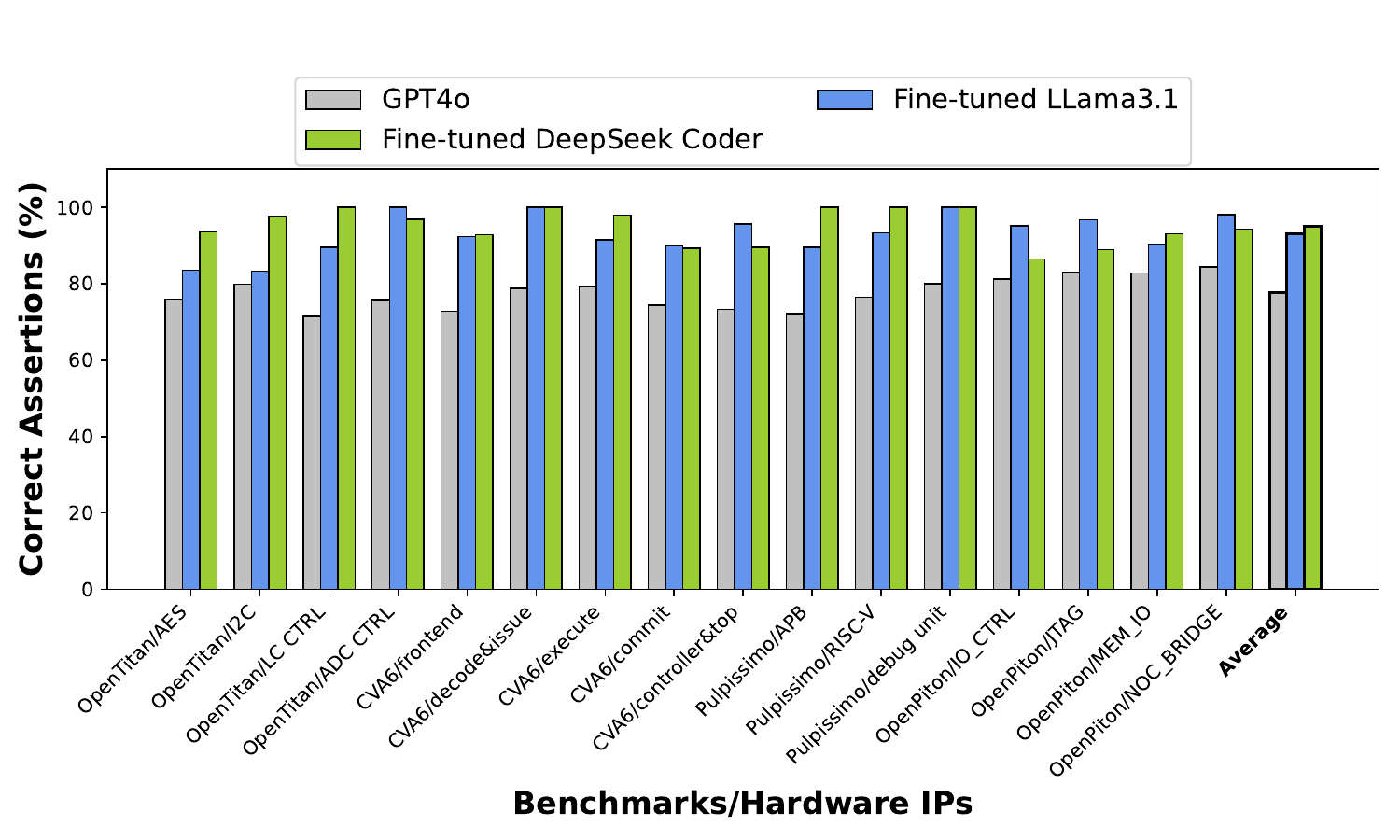}
        \caption{Comparison of GPT-4o and Fine-tuned model in Functionally correct assertions.}
        \label{fig:gpt4_comp_fun}
    \end{subfigure}
    \caption{ Comparison of GPT-4o and Fine-Tuned Model Performance.}
    \label{fig:gpt4_Comp}
\end{figure}

To illustrate \dataset's effectiveness, we compare fine-tuned versions of the DeepSeek Coder and Llama 3.1 model with GPT-4o. Figure \ref{fig:gpt4_comp_syn} and Figure \ref{fig:gpt4_comp_fun} show the syntactic and functional correctness of assertions generated by GPT-4o and the Fine-Tuned Llama 3.1 and Deepseek Coder models across various hardware benchmarks. The X-axis represents the benchmark SoC with its corresponding IP (such as OpenTitan AES, OpenTitan I2C, OpenTitan LC CTRL, and CVA6/Frontend), where assertions are evaluated. The Y-axis displays the percentage of correct assertions, indicating how reliably each model generated the assertions. Figure \ref{fig:gpt4_comp_syn} shows that both fine-tuned Llama 3.1 and Deepseek Coder models significantly outperform GPT-4o by up to 20.69\% in generating syntactically correct assertions. Similarly, as evident in Figure \ref{fig:gpt4_comp_fun}, both Llama 3.1 and Deepseek Coder outperformed GPT-4o by as much as 24.14\% and 21.02\% respectively, with functionally correct assertion in modules such as CVA6/Decode\&Issue and Pulpissimo/Debug unit. These results emphasize that LLMs fine-tuned on \dataset enhance not only syntactic correctness but also the functional reliability of the generated hardware assertions.

\subsubsection{Coverage Measurement:}
\label{sec:coverage_measurement_appen}
In our approach, we use Complete Path Coverage as our primary coverage metric. This ensures a comprehensive evaluation of the system's behavior by accounting for all potential paths. To validate our coverage, we employed both formal and simulation-based verification tools, including Cadence JasperGold and Xilinx Vivado. These tools allowed us to rigorously analyze the generated assertions and ensure that they comprehensively cover all the functions defined within the system. By leveraging our method to extract properties from every possible conditional branch, we achieve up to 100\% CPC for both Llama 3.1 and Deepseek Coder. This verification strategy confirms the correctness and reliability of the design's functionality across all defined behaviors.

\subsection{Ablation Study with Uncleaned Variable Names}
\label{sec:abl_study}

\input{Tables/abl_table}
Tables~\ref{tab:abl_table_syntactic} and~\ref{tab:abl_table_functional} present an ablation study with uncleaned variable names to highlight the impact of cleaning variable names on both syntactical and functional correctness. Table~\ref{tab:abl_table_syntactic} reports the results for Syntactically Correct Assertions, while Table~\ref{tab:abl_table_functional} shows the results for Functionally Correct Assertions. In each table, Column~3 corresponds to assertions generated by the base model. Column~4 presents assertions generated by a model fine-tuned on a dataset containing syntactically incorrect variables (\textit{e.g.,} special characters not allowed in HDL languages). Column~5 shows results from a model fine-tuned on a dataset with duplicate variables that may skew the model’s learning and introduce ambiguity, and Column~6 reports results from a model fine-tuned on a dataset with inconsistent variables (such as conflicting variable names). Finally, Column~7 in each table reports the results using the cleaned variables that we ultimately use to build VERT.

The results reveal a clear trend across both syntactically and functionally correct assertions: cleaning variable names leads to significant improvements in performance. For instance, in Table~\ref{tab:abl_table_syntactic}, the base model for Llama~3.1 achieves only 34.65\% syntactically correct assertions on average, and this performance does not improve when the model is fine-tuned on data containing syntactically incorrect variables (dropping slightly to 32.89\%). In contrast, when the model is trained on data with duplicate variables, the average correctness nearly doubles to 77.66\%. This improvement continues with inconsistent variables (up to 91.18\%), and ultimately, the highest performance of 94.21\% is reached once the variables are cleaned. A similar pattern holds for DeepSeek Coder, where the average syntactic correctness increases from around 23\% (base or syntactically incorrect) to 80.11\% with duplicate variables, 93.93\% with inconsistent variables, and peaks at 96.52\% with cleaned variables.

Table~\ref{tab:abl_table_functional} shows identical trends for functionally correct assertions. Llama~3.1’s base performance is at 22.50\%, with negligible change when trained on syntactically incorrect variables (21.40\%). However, exposure to duplicate variables boosts the average to 77.00\%, while training on inconsistent variables increases it further to 90.18\%. Cleaning the variables finally results in a remarkable performance of about 93.89\% on average, indicating that proper variable naming is crucial not just for syntactical correctness but also for ensuring functional correctness. DeepSeek Coder exhibits a comparable progression, improving from a base of 19.91\% to 96.25\% with cleaned variables.

Notably, certain hardware modules highlight these effects even more starkly. For example,  for syntactically correct assertions, Llama~3.1’s performance on the OpenTitan/ADC CTRL design jumps from 17.46\% (base) to 100\% with cleaned variables, and similar improvements are observed across various IPs. These results confirm that uncleaned or inconsistent variable names severely hinder the model’s ability to generate correct assertions. In contrast, cleaning the variable names not only rectifies the syntactical issues but also enables the models to generate assertions that are functionally more accurate, thereby validating our approach outlined in Section~\ref{sec:DatasetComposition}.

\input{Appendix_syn_func/Keypoint_abl_stdy}

\subsection{Examples of LLMs Fine-tuned on VERT Addressing GPT-4o Issues and Comparison With Human Experts}
\label{sec:label}
In this section, we show how the intuition in dataset formulation, as discussed in Section 4.1, translates into the effectiveness of the assertions generated by LLMs fine-tuned on VERT. Furthermore, we will compare these models with those of human experts. These scenarios include (1) clock cycle and pre-condition capturing, (2) relation representation for nested conditional blocks, and (3) handling long conditions. Each section begins with the relevant code snippet, followed by the assertions generated by each approach.

\subsubsection{ LLMs Fine-tuned on VERT vs. GPT-4o:\newline} 
\label{vertvsgpt4}

\textbf{Clock Cycle Misinterpretations:} We observe that LLMs often get confused between using the overlapping implication symbol ($|->$) and the non-overlapping symbol ($|=>$). To resolve clock cycle misinterpretation by LLMs, we standardized our format by using the overlapping implication symbol with a specified delay count and replacing the non-overlapping symbol. As shown in Textbox \ref{tbox:sv_clkcycle_appn}, GPT-4o incorrectly uses the non-overlapping symbol, but both LLMs fine-tuned on VERT address this issue by correctly interpreting the clock cycle by using the overlapping implication symbol.
\input{Appendix_exmaples_Addressing_Issues/Clock_cycle}

\textbf{Miscapturing if Condition for else Branches:} LLMs often overlook the previous if/else if conditions when generating assertions for the subsequent else or else if branches in conditional statements. VERT addresses the common omission of conditions in the else/else-if branches of if-else statements by exposing the model to diverse conditional structures. As shown in Textbox \ref{tbox:sv_nestedIF_appn},  GPT-4o, in its second assertion, misses that the second assertion should verify that $monitor\_enable$ is \textit{FALSE} and $monitor\_enable\_q$ is \textit{TRUE}. However, both LLMs fine-tuned on VERT correctly identify these conditions and include them in the fourth generated assertion.

\textbf{Nested If-Else Conditions:} LLMs often struggle with generating assertions for deeply nested if-else conditions (\textit{i.e.,} blocks nested beyond two levels), which are common in hardware design. To address the challenge of LLMs struggling with deeply nested if-else statements, we expanded VERT to include complex, multi-level conditional structures. As illustrated by Textbox \ref{tbox:sv_nestedIF_appn}, GPT-4o incorrectly represents the logic of a nested block. Specifically, it ignores the if condition and creates an assertion using a ternary operator, where the inner if condition is used as the test, the assignment is the \textit{TRUE} branch, and \textit{1'b1} is the \textit{TRUE} branch. However, both LLMs fine-tuned on VERT to correctly identify the correct nested statement and their conditions.

\textbf{Handling Long Conditions:} To address the challenge of generating accurate assertions for long and complex conditions, we expanded VERT to include a variety of cases where multiple conditions and operators must be evaluated simultaneously. As shown in textbox \ref{tbox:sv_long_appn}, the source code requires any one of six conditions to be met before raising the $mr\_err$ flag. However, the generated output by GPT-4o creates three assertions for one flag change, each neglecting the other three conditions. This not only leads to incorrect assertions but also causes overhead in the number of assertions. However, both LLMs fine-tuned on VERT correctly generate a singular assertion that describes all the conditions and their relationships within the if-statement.

\input{Appendix_exmaples_Addressing_Issues/nested_if}

\input{Appendix_exmaples_Addressing_Issues/Long_Conditions}

\subsubsection{ LLMs Fine-tuned on VERT vs. Human Expert:\newline} 

Human experts and LLMs fine-tuned on \dataset generate assertions differently. Experts often create simpler, more readable assertions, reflecting a preference for human-friendly formats. LLMs, on the other hand, produce more standardized logical expressions. This difference highlights style preferences rather than superiority. Notably, despite these stylistic differences, LLMs fine-tuned on \dataset achieve assertion generation performance comparable to that of human experts. This is illustrated in the Textbox \ref{tbox:hum_exp}. 
For instance, in Example 1, when verifying whether the \texttt{bus\_release\_cnt} signal has been reset, the LLM-generated assertion takes the form \texttt{bus\_release\_cnt == `0'}, while the human expert expresses the same operation as \texttt{!(|bus\_release\_cnt)}. The latter format, often favored by experts, is typically chosen for its ease of writing and brevity. Similarly, in Example 2, since \textit{mr\_err} is a one-bit flag, the human expert chooses to verify whether the flag was raised by writing \texttt{mr\_err} for the check, while the LLMs use the full expression, \texttt{mr\_err == 1'b1}.


\input{Appendix_exmaples_Addressing_Issues/HumanExpert}

\input{Appendix_syn_func/Keypoint_quail}

\subsection{Impact of Contamination on Assertion Generation}
\label{sec:cont}

Contamination refers to the presence of extraneous HDL components in the design files that can interfere with the assertion generation process. In particular, module instantiations and ``\texttt{ifdef}'' commands act as contaminants, hindering the creation of syntactically and functionally correct assertions. This issue is evident in the models used in Section~\ref{Sec:EvalResults}, which rarely generate assertions from module instantiations. This often results in syntactically and functionally incorrect outputs. Moreover, these smaller models tend to misinterpret ``\texttt{ifdef}'' commands as conventional if-else statements. While this misclassification occurs infrequently, it reduces the percentage of correctly generated assertions. Assertions generated from these commands are often incorrect syntactically and inadequate functionally. They are syntactically flawed because \texttt{ifdef} commands do not follow standard if-else syntax and lack the necessary details for proper branching. Functionally, they are inadequate because they do not contribute effectively to branching logic. 



\input{Tables/contamination}

Table~\ref{tab:cont-table} illustrates the effect of increasing contamination in design files on assertion generation. In this context, contamination refers to the addition of extra ``\texttt{ifdef}'' commands and module instantiations. For example, “+10 contamination” means that 10 additional instances of each element were introduced beyond those originally present.

The table’s first column specifies the LLM being tested, while Column 2 identifies the specific test benchmark used. Columns 3 through 5 display the total number of assertions generated under three conditions: no contamination, an additional 10 contamination, and an additional 20 contamination. Columns 6 through 8 indicate the percentage of assertions that are syntactically valid at each contamination level. Finally, Columns 9 through 11, labeled report the percentage of assertions that are logically accurate and align with the intended functionality.

A contamination level below 10 was found to have a negligible impact, whereas levels exceeding 20 proved impractical due to the context size limitations of our models. Overall, as the contamination level increases, the number of incorrectly generated assertions also rises, resulting in a 3\% drop in accuracy. It should be noted that typically, in the hardware design, the number of ``\texttt{ifdef}'' commands is limited to at most five. \textbf{As a result, the scenarios used to evaluate the contamination effect represent the worst-case scenarios.} Although these conditions are unrealistic for typical applications, they provide a stringent framework for studying the impact of contamination on the models.

%% file: Appendix_syn_func/textbox.tex
\begin{mybox}[colback=white, colframe=black, boxrule=0.2mm, sharp corners, coltitle=black, colbacktitle=white, label=tb:diff_func_syn ]{Syntactical and Functional Errors in Assertions.}
\lstset{style=mystyle2}

Design Source Code:

\begin{lstlisting}[language=Verilog]
always_ff @(posedge clk_aon_i or negedge rst_aon_ni) begin
    if (!rst_aon_ni) begin
        wakeup_timer_cnt_q    <= '0;
    end
    else if (wakeup_timer_cnt_clr || cfg_fsm_rst_i || trigger_h2l) begin
        wakeup_timer_cnt_q <= '0;
    end else begin
        wakeup_timer_cnt_q <= wakeup_timer_cnt_d;
    end
end
\end{lstlisting}

Here is the requested property:
\begin{lstlisting}[language=Verilog]
property ResetTimer1; @(posedge clk_aon_i) (!rst_aon_ni)|->wakeup_timer_cnt_q=1'b0; endproperty 
// A syntactically incorrect assertion
// Here, instead of using `==' symbol, `=' was used
property ResetTimer2;
@(posedge clk_aon_i)(wakeup_timer_cnt_clr||cfg_fsm_rst_i||trigger_h2l)|->wakeup_timer_cnt_q=='0;
endproperty
// A Functionally incorrect assertion
// Here, the generated assertion is missed capturing the `if' condition.
\end{lstlisting}

\end{mybox}


%% file: Appendix_syn_func/keypoint.tex



\begin{fncyBox}
    \textbf{Key Point:} We highlight how both types of errors (syntactical and functional) can undermine the correctness of assertions, either by preventing successful compilation (in the case of syntax errors) or by failing to accurately verify the design's intended behavior (in the case of functional errors).
\end{fncyBox}


%% file: Tables/LeakageTable.tex
\begin{table}[b]
\caption{Overlap scores between \dataset and the hardware IP benchmarks used for evaluation.}
\label{tab:LeakageTable}
\resizebox{\textwidth}{!}{%
\renewcommand{\arraystretch}{1.5}
\begin{tabular}{l|c|c|c|c}
\hline
\multicolumn{1}{c|}{\textbf{Benchmark/ Hardware IP}} & \textbf{Code Overlap Score} & \textbf{Assertions Overlap Score} & \textbf{\begin{tabular}[c]{@{}c@{}}DeepSeek Coder Prompt \\ Overlap Score\end{tabular}} & \textbf{\begin{tabular}[c]{@{}c@{}}Llama 3.1 Prompt \\ Overlap Score\end{tabular}} \\ \hline
OpenTitan/AES                                        & 0.0003                      & 0.0000                            & 0.0004                                                                                  & 0.0004                                                                             \\
OpenTitan/I2C                                        & 0.0003                      & 0.0000                            & 0.0006                                                                                  & 0.0006                                                                             \\
OpenTitan/LC CTRL                                    & 0.0002                      & 0.0000                            & 0.0005                                                                                  & 0.0005                                                                             \\
OpenTitan/ADC CTRL                                   & 0.0002                      & 0.0000                            & 0.0005                                                                                  & 0.0005                                                                             \\ \hline
CVA6/Frontend                                        & 0.0003                      & 0.0000                            & 0.0005                                                                                  & 0.0004                                                                             \\
CVA6/Decode\&Issue                                   & 0.0003                      & 0.0000                            & 0.0004                                                                                  & 0.0004                                                                             \\
CVA6/Execute                                         & 0.0002                      & 0.0000                            & 0.0004                                                                                  & 0.0004                                                                             \\
CVA6/Commit                                          & 0.0002                      & 0.0000                            & 0.0004                                                                                  & 0.0004                                                                             \\
CVA6/Controller\&Top                                 & 0.0001                      & 0.0000                            & 0.0002                                                                                  & 0.0002                                                                             \\ \hline
Pulpissimo/APB                                       & 0.0002                      & 0.0000                            & 0.0004                                                                                  & 0.0004                                                                             \\
Pulpissimo/RISCV                                     & 0.0005                      & 0.0000                            & 0.0005                                                                                  & 0.0004                                                                             \\
Pulpissimo/debug\_unit                               & 0.0002                      & 0.0000                            & 0.0004                                                                                  & 0.0005                                                                             \\ \hline

OpenPiton/IO\_CTRL                                   & 0.0001                      & 0.0000                            & 0.0002                                                                                  & 0.0002                                                                              \\
OpenPiton/JTAG                                       & 0.0002                      & 0.0000                            & 0.0004                                                                                  & 0.0004                                                                              \\
OpenPiton/MEM\_IO                                    & 0.0003                      & 0.0000                            & 0.0003                                                                                 & 0.0003                                                                              \\
OpenPiton/NOC\_BRIDGE                                & 0.0002                      & 0.0000                            & 0.0007                                                                                  & 0.0007                                                                              \\ \hline
\end{tabular}%
}
\end{table}


%% file: Appendix_syn_func/keypoint_overlap.tex
\begin{fncyBox}
    \textbf{Key Point:} The byte-level 13-gram analysis confirms that VERT has negligible content overlap with the evaluation benchmarks, ensuring that performance assessments remain unbiased and truly reflective of model generalization.
\end{fncyBox}

%% file: Tables/hyperparam.tex
\begin{wraptable}{r}{0.45\textwidth}
\vspace{-10mm}
\caption{Training Hyperparameters}
\vspace{-3mm}
\newcolumntype{M}[1]{>{\centering\arraybackslash}m{#1}}
\begin{tabular}{|l|c|}
\hline
\rowcolor[HTML]{c0c0c0} 
\multicolumn{1}{|c|}{\textbf{Hyperparameter}} & \textbf{Value} \\ \hline \hline
Lora RANK                & 256            \\ \hline
Lora alpha              & 256            \\ \hline
Maximum sequence length  & 4096           \\ \hline
Epochs                   & 3              \\ \hline
Batch                    & 64             \\ \hline
Learning\_rate           & 1.00$e^{-04}$  \\ \hline
Training Presicion       & Bf16           \\ \hline
\end{tabular}%
\label{tab:param}
\vspace{-4mm}
\end{wraptable} 


%% file: Tables/asserttable.tex
\begin{table*}[t]
\centering
\caption{Performance Comparison of base and fine-tuned models on assertion generation across various hardware IP benchmarks. }
\label{tab:asserttable}
\resizebox{\textwidth}{!}{%
\renewcommand{\arraystretch}{1.5}
\begin{tabular}{l|l|cc|cc|cc}
\hline
\multirow{2}{*}{\textbf{Models}} &
  \multicolumn{1}{c|}{\multirow{2}{*}{\begin{tabular}[c]{@{}c@{}}\textbf{Benchmark/}\\ \textbf{Hardware IP}\end{tabular}}} &
  \multicolumn{2}{c|}{\textbf{Generated Assertions}} &
  \multicolumn{2}{c|}{\textbf{Syntactically Correct Assertions} (\%)} &
  \multicolumn{2}{c}{\textbf{Functionally Correct Assertions} (\%)} \\
 &
  \multicolumn{1}{c|}{} &
  \multicolumn{1}{l|}{Base Model} &
  \multicolumn{1}{l|}{Fine-Tuned Model} &
  \multicolumn{1}{l|}{Base Model} &
  \multicolumn{1}{l|}{Fine-Tuned Model} &
  \multicolumn{1}{l|}{Base Model} &
  \multicolumn{1}{l}{Fine-Tuned Model} \\ \hline
\multirow{16}{*}{\textbf{Llama 3.1}}     & OpenTitan/AES                & 212 & 125 & 35.84 & 88.70  & 8.02  & 83.48  \\
                                & OpenTitan/I2C                & 149 & 126 & 29.53 & 83.33  & 9.39  & 83.33  \\
                                & OpenTitan/LC CTRL            & 26  & 19  & 23.07 & 89.47  & 7.69  & 89.47  \\
                                & OpenTitan/ADC CTRL           & 63  & 32  & 17.46 & 100.00 & 9.52  & 100.00 \\ \cline{2-8} 
                                & CVA6/Frontend                & 17  & 13  & 41.18 & 92.31  & 11.76 & 92.31  \\
                                & CVA6/Decode\&Issue           & 31  & 34  & 22.58 & 100.00 & 6.45  & 100.00 \\
                                & CVA6/Execute                 & 110 & 105 & 25.55 & 91.43  & 5.45  & 91.43  \\
                                & CVA6/Commit                  & 70  & 79  & 38.57 & 89.87  & 10    & 89.87  \\
                                & CVA6/Controller\&Top         & 73  & 68  & 34.24 & 95.59  & 5.48  & 95.59  \\ \cline{2-8} 
                                & Pulpissimo/APB               & 15  & 19  & 53.33 & 89.47  & 53.33 & 89.47  \\
                                & Pulpissimo/RISCV & 19  & 15  & 21.05 & 93.33  & 21.05 & 93.33  \\
                                & Pulpissimo/debug\_unit        & 6   & 11  & 16.67 & 100.00 & 16.67 & 100.00 \\ \cline{2-8}
                                & OpenPiton/IO\_CTRL & 136 & 124 & 45.58 & 99.19 & 13.24 & 95.16 \\
                                & OpenPiton/JTAG & 47 & 30 & 48.93 & 96.67 & 19.15 & 96.67 \\
                                & OpenPiton/MEM\_IO & 68 & 52 & 61.76 & 100.00 & 20.59 & 90.38 \\
                                & OpenPiton/NOC\_BRIDGE & 59 & 52 & 38.98 & 98.07 & 18.65 & 98.07 \\ \hline

\multirow{16}{*}{\textbf{DeepSeek Coder}} & OpenTitan/AES                & 148 & 157 & 10.81 & 94.90  & 6.08  & 93.63  \\
                                & OpenTitan/I2C                & 132 & 124 & 12.12 & 97.58  & 8.33  & 97.58  \\
                                & OpenTitan/LC CTRL            & 21  & 19  & 14.25 & 100.00 & 9.52  & 100.00 \\
                                & OpenTitan/ADC CTRL           & 32  & 32  & 6.25  & 100.00 & 0     & 96.88  \\ \cline{2-8} 
                                & CVA6/Frontend                & 16  & 14  & 56.25 & 92.86  & 37.5  & 92.86  \\
                                & CVA6/Decode\&Issue           & 37  & 32  & 18.92 & 100.00 & 13.51 & 100.00 \\
                                & CVA6/Execute                 & 91  & 99  & 26.37 & 97.98  & 20.88 & 97.98  \\
                                & CVA6/Commit                  & 97  & 93  & 21.65 & 89.25  & 17.53 & 89.25  \\
                                & CVA6/Controller\&Top         & 82  & 76  & 21.95 & 89.47  & 15.85 & 89.47  \\ \cline{2-8} 
                                & Pulpissimo/APB               & 25  & 19  & 24.00 & 100.00 & 24.00 & 100.00 \\
                                & Pulpissimo/RISCV & 13  & 15  & 23.08 & 100.00 & 23.08 & 100.00 \\
                                & Pulpissimo/debug\_unit        & 11  & 11  & 15.38 & 100.00 & 15.38 & 100.00 \\ \cline{2-8}
                                & OpenPiton/IO\_CTRL & 147 & 103 & 26.53 & 92.23 & 9.52 & 86.41 \\
                                & OpenPiton/JTAG & 43 & 27 & 32.56 & 100.00 & 18.60 & 88.88 \\
                                & OpenPiton/MEM\_IO & 82 & 72 & 32.93 & 95.83 & 12.19 & 93.05 \\
                                & OpenPiton/NOC\_BRIDGE & 66 & 35 & 34.85 & 94.28 & 13.64 & 94.28 \\ \hline
\end{tabular}%
}
\end{table*}

%% file: Tables/abl_table.tex
\begin{table}[h]
\centering
\caption{Ablation Study with Uncleaned Variable Names -- Syntactically Correct Assertions (\%)}
\label{tab:abl_table_syntactic}
\renewcommand{\arraystretch}{1.5}
\resizebox{\textwidth}{!}{%
  \newcolumntype{M}[1]{>{\centering\arraybackslash}m{#1}}
  \newcolumntype{R}{>{\Large\centering\arraybackslash}c}
  \begin{tabular}{c|c|*{5}{R}}
    \hline
    \multirow{2}{*}{\textbf{Models}} & 
    \multirow{2}{*}{\textbf{\begin{tabular}[c]{@{}c@{}}Benchmark/\\ Hardware IP\end{tabular}}} & 
    \multicolumn{5}{c}{\textbf{Syntactically Correct Assertions (\%)}} \\
    & & 
    \textbf{Base Model} & 
    \textbf{\begin{tabular}[c]{@{}c@{}}With Syntactically \\ Incorrect Variables\end{tabular}} & 
    \textbf{\begin{tabular}[c]{@{}c@{}}With Duplicate \\ Variables\end{tabular}} & 
    \textbf{\begin{tabular}[c]{@{}c@{}}With Inconsistent \\ Variables\end{tabular}} & 
    \textbf{\begin{tabular}[c]{@{}c@{}}Cleaned \\ Variables\end{tabular}} \\ \hline
    \multirow{17}{*}{\textbf{Llama 3.1}} & 
      OpenTitan/AES         & 35.84 & 35.20 & 72.80 & 86.40 & 88.70 \\
    & OpenTitan/I2C         & 29.53 & 28.57 & 66.67 & 83.33 & 83.33 \\
    & OpenTitan/LC CTRL     & 23.07 & 21.05 & 73.68 & 84.21 & 89.47 \\
    & OpenTitan/ADC CTRL    & 17.46 & 18.75 & 81.25 & 90.63 & 100.00 \\ \cline{2-7} 
    & CVA6/Frontend         & 41.18 & 38.46 & 76.92 & 92.31 & 92.31 \\
    & CVA6/Decode\&Issue     & 22.58 & 23.53 & 82.35 & 94.12 & 100.00 \\
    & CVA6/Execute          & 25.55 & 24.76 & 74.29 & 85.71 & 91.43 \\
    & CVA6/Commit           & 38.57 & 35.44 & 73.42 & 88.61 & 89.87 \\
    & CVA6/Controller\&Top   & 34.24 & 32.35 & 79.41 & 95.59 & 95.59 \\ \cline{2-7} 
    & Pulpissimo/APB        & 53.33 & 52.63 & 73.68 & 89.47 & 89.47 \\
    & Pulpissimo/RISCV      & 21.05 & 20.00 & 80.00 & 93.33 & 93.33 \\
    & Pulpissimo/debug\_unit & 16.67 & 18.18 & 81.82 & 90.91 & 100.00 \\ \cline{2-7}
    & OpenPiton/IO\_CTRL    & 45.58 & 46.38 & 78.5  & 96.77 & 99.19 \\
    & OpenPiton/JTAG        & 48.93 & 46.8  & 82.66 & 93.33 & 96.67 \\
    & OpenPiton/MEM\_IO     & 61.76 & 53.83 & 91.32 & 100   & 100   \\
    & OpenPiton/NOC\_BRIDGE & 38.98 & 30.18 & 73.86 & 94.23 & 98.07 \\ \cline{2-7}
    & \textbf{Average}      & \textbf{34.65} & \textbf{32.88} & \textbf{77.66} & \textbf{91.18} & \textbf{94.21} \\ \hline
    \multirow{17}{*}{\textbf{DeepSeek Coder}} & 
      OpenTitan/AES         & 10.81 & 10.19 & 75.80 & 89.17 & 94.90 \\
    & OpenTitan/I2C         & 12.12 & 11.29 & 79.03 & 95.16 & 97.58 \\
    & OpenTitan/LC CTRL     & 14.25 & 15.79 & 84.21 & 94.74 & 100.00 \\
    & OpenTitan/ADC CTRL    & 6.25  & 6.25  & 78.13 & 96.88 & 100.00 \\ \cline{2-7} 
    & CVA6/Frontend         & 56.25 & 57.14 & 78.57 & 92.86 & 92.86 \\
    & CVA6/Decode\&Issue     & 18.92 & 18.75 & 84.38 & 96.88 & 100.00 \\
    & CVA6/Execute          & 26.37 & 25.25 & 77.78 & 97.98 & 97.98 \\
    & CVA6/Commit           & 21.65 & 20.43 & 75.27 & 82.80 & 89.25 \\
    & CVA6/Controller\&Top   & 21.95 & 21.05 & 75.00 & 86.84 & 89.47 \\ \cline{2-7} 
    & Pulpissimo/APB        & 24    & 26.32 & 78.95 & 94.74 & 100.00 \\
    & Pulpissimo/RISCV      & 23.08 & 20.00 & 80.00 & 100.00& 100.00 \\
    & Pulpissimo/debug\_unit & 15.38 & 18.18 & 81.82 & 100.00& 100.00 \\ \cline{2-7}
    & OpenPiton/IO\_CTRL    & 26.53 & 24.32 & 68.38 & 90.29 & 92.23 \\
    & OpenPiton/JTAG        & 32.56 & 35.24 & 93.12 & 100   & 100   \\
    & OpenPiton/MEM\_IO     & 32.93 & 33.26 & 88.76 & 93.05 & 95.83 \\
    & OpenPiton/NOC\_BRIDGE & 34.85 & 36.3  & 82.54 & 91.43 & 94.28 \\ \cline{2-7}
    & \textbf{Average}      & \textbf{23.62} & \textbf{23.74} & \textbf{80.11} & \textbf{93.93} & \textbf{96.52} \\ \hline
  \end{tabular}%
}
\end{table}

\begin{table}[h]
\centering
\caption{Ablation Study with Uncleaned Variable Names -- Functionally Correct Assertions (\%)}
\label{tab:abl_table_functional}
\renewcommand{\arraystretch}{1.5}
\resizebox{\textwidth}{!}{%
  \newcolumntype{M}[1]{>{\centering\arraybackslash}m{#1}}
  \newcolumntype{R}{>{\Large\centering\arraybackslash}c}
  \begin{tabular}{c|c|*{5}{R}}
    \hline
    \multirow{2}{*}{\textbf{Models}} & 
    \multirow{2}{*}{\textbf{\begin{tabular}[c]{@{}c@{}}Benchmark/\\ Hardware IP\end{tabular}}} & 
    \multicolumn{5}{c}{\textbf{Functionally Correct Assertions (\%)}} \\
    & & 
    \textbf{Base Model} & 
    \textbf{\begin{tabular}[c]{@{}c@{}}With Syntactically \\ Incorrect Variables\end{tabular}} & 
    \textbf{\begin{tabular}[c]{@{}c@{}}With Duplicate \\ Variables\end{tabular}} & 
    \textbf{\begin{tabular}[c]{@{}c@{}}With Inconsistent \\ Variables\end{tabular}} & 
    \textbf{\begin{tabular}[c]{@{}c@{}}Cleaned \\ Variables\end{tabular}} \\ \hline
    \multirow{17}{*}{\textbf{Llama 3.1}} & 
      OpenTitan/AES         & 8.02  & 7.20  & 68.80 & 82.40 & 83.48 \\
    & OpenTitan/I2C         & 9.39  & 9.52  & 66.67 & 80.16 & 83.33 \\
    & OpenTitan/LC CTRL     & 7.69  & 5.26  & 73.68 & 84.21 & 89.47 \\
    & OpenTitan/ADC CTRL    & 9.52  & 9.38  & 81.25 & 96.88 & 100.00 \\ \cline{2-7} 
    & CVA6/Frontend         & 11.76 & 15.38 & 76.92 & 84.62 & 92.31 \\
    & CVA6/Decode\&Issue     & 6.45  & 5.88  & 82.35 & 97.06 & 100.00 \\
    & CVA6/Execute          & 5.45  & 5.71  & 74.29 & 86.67 & 91.43 \\
    & CVA6/Commit           & 10.00 & 10.13 & 73.42 & 89.87 & 89.87 \\
    & CVA6/Controller\&Top   & 5.48  & 5.88  & 79.41 & 89.71 & 95.59 \\ \cline{2-7} 
    & Pulpissimo/APB        & 53.33 & 52.63 & 73.68 & 89.47 & 89.47 \\
    & Pulpissimo/RISCV      & 21.05 & 20.00 & 73.33 & 86.67 & 93.33 \\
    & Pulpissimo/debug\_unit & 16.67 & 18.18 & 81.82 & 90.91 & 100.00 \\ \cline{2-7}
    & OpenPiton/IO\_CTRL    & 45.58 & 46.38 & 78.5  & 96.77 & 99.19 \\
   & OpenPiton/JTAG        & 48.93 & 46.8  & 82.66 & 93.33 & 96.67 \\
   & OpenPiton/MEM\_IO     & 61.76 & 53.83 & 91.32 & 100   & 100   \\
   & OpenPiton/NOC\_BRIDGE & 38.98 & 30.18 & 73.86 & 94.23 & 98.07 \\ \cline{2-7}
    & \textbf{Average}      & \textbf{22.50} & \textbf{21.40} & \textbf{77.00} & \textbf{90.18} & \textbf{93.89} \\ \hline
    \multirow{17}{*}{\textbf{DeepSeek Coder}} & 
      OpenTitan/AES         & 6.08  & 6.37  & 75.80 & 92.36 & 93.63 \\
    & OpenTitan/I2C         & 8.33  & 8.06  & 79.03 & 95.16 & 97.58 \\
    & OpenTitan/LC CTRL     & 9.52  & 10.53 & 84.21 & 94.74 & 100.00 \\
    & OpenTitan/ADC CTRL    & 0.00  & 0.00  & 78.13 & 93.75 & 96.88 \\ \cline{2-7} 
    & CVA6/Frontend         & 37.50 & 35.71 & 78.57 & 85.71 & 92.86 \\
    & CVA6/Decode\&Issue     & 13.51 & 12.50 & 84.38 & 93.75 & 100.00 \\
    & CVA6/Execute          & 20.88 & 20.20 & 77.78 & 92.93 & 97.98 \\
    & CVA6/Commit           & 17.53 & 17.20 & 75.27 & 81.72 & 89.25 \\
    & CVA6/Controller\&Top   & 15.85 & 14.47 & 75.00 & 85.53 & 89.47 \\ \cline{2-7} 
    & Pulpissimo/APB        & 24.00 & 26.32 & 78.95 & 100.00& 100.00 \\
    & Pulpissimo/RISCV      & 23.08 & 20.00 & 80.00 & 93.33 & 100.00 \\
    & Pulpissimo/debug\_unit & 15.38 & 18.18 & 81.82 & 90.91 & 100.00 \\ \cline{2-7}
    & OpenPiton/IO\_CTRL    & 26.53 & 24.32 & 68.38 & 90.29 & 92.23 \\
   & OpenPiton/JTAG        & 32.56 & 35.24 & 93.12 & 100   & 100   \\
   & OpenPiton/MEM\_IO     & 32.93 & 33.26 & 88.76 & 93.05 & 95.83 \\
   & OpenPiton/NOC\_BRIDGE & 34.85 & 36.3  & 82.54 & 91.43 & 94.28 \\ \cline{2-7}
    & \textbf{Average}      & \textbf{19.91} & \textbf{19.92} & \textbf{80.11} & \textbf{92.17} & \textbf{96.25} \\ \hline
  \end{tabular}%
}
\end{table}

%% file: Appendix_syn_func/Keypoint_abl_stdy.tex
\begin{fncyBox}
    \textbf{Key Point:} The ablation study demonstrates that cleaning and standardizing variable names is crucial for improving both the syntactical and functional correctness of generated assertions, thereby enhancing overall model reliability.
\end{fncyBox}

%% file: Appendix_exmaples_Addressing_Issues/Clock_cycle.tex
\begin{mybox}[colback=white, colframe=black, boxrule=0.2mm, sharp corners, coltitle=black, colbacktitle=white, label=tbox:sv_clkcycle_appn]{Clock Cycle and Pre-condition Capturing in GPT-4o VS Vs LLMs fine tuned on VERT.}
Generate a list of synchronous SystemVerilog assertions executing at \texttt{(\textcolor{blue}{\textit{posedge}} clk\_i)} from the following code:
\lstset{style=mystyle2}
\begin{lstlisting}[language=Verilog][h]
always @(posedge clk_i or negedge rst_ni) begin
    if (~rst_ni) begin
        wr_state_q <= IDLE;
        wr_cnt_q   <= '0;
    end else begin
        wr_state_q <= wr_state_d;
        wr_cnt_q   <= wr_cnt_d;
    end
end
\end{lstlisting}

GPT-4o Response:

\begin{lstlisting}[language=Verilog]
...
// Assertion to ensure that wr_state_q follows wr_state_d when rst_ni is high
assert property (@(posedge clk_i) (rst_ni && (wr_state_d !== IDLE)) |=> (wr_state_q == wr_state_d))
  else $error("State transition failure: wr_state_q did not follow wr_state_d.");

// Assertion to ensure that wr_cnt_q follows wr_cnt_d when rst_ni is high
assert property (@(posedge clk_i) (rst_ni && (wr_cnt_d !== '0)) |=> (wr_cnt_q == wr_cnt_d))
  else $error("Counter transition failure: wr_cnt_q did not follow wr_cnt_d.");

// Assertion to check that the state and count do not update during reset
assert property (@(posedge clk_i) (~rst_ni) |=> (wr_state_q == IDLE && wr_cnt_q == '0))
  else $error("Reset condition not maintained: wr_state_q or wr_cnt_q changed during reset.");
...
\end{lstlisting}

Llama 3.1 fine-tuned on \dataset Response:

\begin{lstlisting}[language=Verilog]

property ResetSynceotid; @(posedge clk_i) ( ~rst_ni ) |-> wr_state_q == IDLE && wr_cnt_q == '0 ; endproperty 
 
property ResetSynceotid; @(posedge clk_i)  ( rst_ni )  |-> wr_state_q == wr_state_d && wr_cnt_q == wr_cnt_d; endproperty 
\end{lstlisting}

DeepSeek Coder fine-tuned on \dataset Response:

\begin{lstlisting}[language=Verilog]
property ResetSynceotid; @(posedge clk_i) ( ~rst_ni ) |-> wr_state_q == IDLE && wr_cnt_q == '0 ; endproperty 
 
property ResetSynceotid; @(posedge clk_i)  ( rst_ni )  |-> wr_state_q == wr_state_d && wr_cnt_q == wr_cnt_d; endproperty 
\end{lstlisting}
\end{mybox}


%% file: Appendix_exmaples_Addressing_Issues/nested_if.tex
\begin{mybox}[colback=white, colframe=black, boxrule=0.2mm, sharp corners, coltitle=black, colbacktitle=white, label=tbox:sv_nestedIF_appn] {Relation Representation for Nested Conditional Blocks in GPT-4o VS Vs LLMs fine tuned on VERT.}
Generate a list of asynchronous SystemVerilog Assertions from the following code:
\lstset{style=mystyle2}
\begin{lstlisting}[language=Verilog]
if (!rst_ni) begin
      bus_release_cnt <= '0;   
end else if (monitor_enable &&!monitor_enable_q) begin
    if (multi_controller_enable_i) begin
        bus_release_cnt <= 30'(bus_inactive_timeout_i);
    end
end else if (bus_release_cnt_load) begin
    bus_release_cnt <= bus_release_cnt_sel;
end else if (bus_release_cnt_dec && (bus_release_cnt!= '0)) begin
    bus_release_cnt <= bus_release_cnt - 1'b1;
end
\end{lstlisting}

GPT-4o Response:

\begin{lstlisting}[language=Verilog]
...
// Synchronous assertion for monitor enable signal transition
assert property (@(posedge clk) disable iff (!rst_ni)
    (monitor_enable && !monitor_enable_q) |-> (multi_controller_enable_i ? 
    (bus_release_cnt == bus_inactive_timeout_i) : 1'b1)
) else $fatal("Bus release count should be set to bus_inactive_timeout_i when monitor_enable rises and multi_controller_enable_i is high.");

// Synchronous assertion for bus_release_cnt_load condition
assert property (@(posedge clk) disable iff (!rst_ni)
    (bus_release_cnt_load) |-> (bus_release_cnt == bus_release_cnt_sel)
) else $fatal("Bus release count should be loaded with bus_release_cnt_sel when bus_release_cnt_load is asserted.");
...
\end{lstlisting}

Llama 3.1 fine-tuned on \dataset  Response:

\begin{lstlisting}[language=Verilog]
 property ResetSynceotid;  ( !rst_ni ) |-> bus_release_cnt == '0 ;endproperty 
 
 property ResetSynceotid; (  rst_ni ) && ( !monitor_enable || monitor_enable_q ) |-> bus_release_cnt == '0 ;endproperty 
 
 property ResetSynceotid;  (  rst_ni ) &&  ( monitor_enable && !monitor_enable_q ) &&  (  multi_controller_enable_i ) |-> bus_release_cnt == 30'(bus_inactive_timeout_i) ;endproperty 
 
 property ResetSynceotid;  (  rst_ni ) &&  ( !monitor_enable || monitor_enable_q ) && (  bus_release_cnt_load ) |-> bus_release_cnt == bus_release_cnt_sel ;endproperty 
 
 property ResetSynceotid; (  rst_ni ) &&  ( !monitor_enable || monitor_enable_q ) && ( !bus_release_cnt_load ) &&  (  bus_release_cnt_dec && (bus_release_cnt != '0)) |-> bus_release_cnt == (bus_release_cnt - 1'b1) ;endproperty 

\end{lstlisting}

DeepSeek Coder fine-tuned on \dataset  Response:

\begin{lstlisting}[language=Verilog]
 property ResetSynceotid;  ( !rst_ni ) |-> bus_release_cnt == '0 ;endproperty 
 
 property ResetSynceotid;  (  rst_ni ) && ( !monitor_enable || monitor_enable_q ) |-> bus_release_cnt == '0 ;endproperty 
 
 property ResetSynceotid;  (  rst_ni ) &&  ( monitor_enable && !monitor_enable_q ) &&  (  multi_controller_enable_i ) |-> bus_release_cnt == 30'(bus_inactive_timeout_i) ;endproperty 
 
 property ResetSynceotid;  (  rst_ni ) &&  ( !monitor_enable || monitor_enable_q ) && (  bus_release_cnt_load ) |-> bus_release_cnt == bus_release_cnt_sel ;endproperty 
 
 property ResetSynceotid;  (  rst_ni ) &&  ( !monitor_enable || monitor_enable_q ) && ( !bus_release_cnt_load ) &&  (  bus_release_cnt_dec && (bus_release_cnt != '0)) |-> bus_release_cnt == (bus_release_cnt - 1'b1) ;endproperty 

\end{lstlisting}

\end{mybox}

%% file: Appendix_exmaples_Addressing_Issues/Long_Conditions.tex
\begin{mybox}[colback=white, colframe=black, boxrule=0.2mm, sharp corners, coltitle=black, colbacktitle=white, label=tbox:sv_long_appn]{Handling Long Conditions in GPT-4o Vs LLMs fine tuned on VERT.}
Generate a list of asynchronous SystemVerilog Assertions from the following code:
\lstset{style=mystyle2}
\begin{lstlisting}[language=Verilog]
if (state_sel_o      != mr_state_sel[i]     ||
      add_rk_sel_o    != mr_add_rk_sel[i]    ||
      key_full_sel_o  != mr_key_full_sel[i]  ||
      key_dec_sel_o   != mr_key_dec_sel[i]   ||
      key_words_sel_o != mr_key_words_sel[i] ||
      round_key_sel_o != mr_round_key_sel[i]) begin
    mr_err = 1'b1;
end
\end{lstlisting}

GPT -4o Response:

\begin{lstlisting}[language=Verilog]
...
// Asynchronous assertion to check key_full_sel_o against mr_key_full_sel[i]
assert property ((key_full_sel_o != mr_key_full_sel[i]) |-> (mr_err == 1'b1))
  else $error("Asynchronous check failure: key_full_sel_o does not match mr_key_full_sel[%0d] and mr_err is not asserted.", i);

// Asynchronous assertion to check key_dec_sel_o against mr_key_dec_sel[i]
assert property ((key_dec_sel_o != mr_key_dec_sel[i]) |-> (mr_err == 1'b1))
  else $error("Asynchronous check failure: key_dec_sel_o does not match mr_key_dec_sel[%0d] and mr_err is not asserted.", i);

// Asynchronous assertion to check key_words_sel_o against mr_key_words_sel[i]
assert property ((key_words_sel_o != mr_key_words_sel[i]) |-> (mr_err == 1'b1))
  else $error("Asynchronous check failure: key_words_sel_o does not match mr_key_words_sel[%0d] and mr_err is not asserted.", i);
...

\end{lstlisting}

Llama 3.1 fine-tuned on \dataset Response:

\begin{lstlisting}[language=Verilog]

property SyncErreotid; (state_sel_o    != mr_state_sel[i]     ||
          add_rk_sel_o   != mr_add_rk_sel[i]    ||
          key_full_sel_o != mr_key_full_sel[i]  ||
          key_dec_sel_o  != mr_key_dec_sel[i]   ||
          key_words_sel_o!= mr_key_words_sel[i] ||
          round_key_sel_o!= mr_round_key_sel[i]) |-> mr_err = 1'b1 ;
endproperty
\end{lstlisting}
DeepSeek Coder fine-tuned on \dataset  Response:

\begin{lstlisting}[language=Verilog]
 property SyncErreotid; (state_sel_o     != mr_state_sel[i]) ||
     (add_rk_sel_o    != mr_add_rk_sel[i]) ||
     (key_full_sel_o  != mr_key_full_sel[i]) ||
     (key_dec_sel_o   != mr_key_dec_sel[i]) ||
     (key_words_sel_o != mr_key_words_sel[i]) ||
     (round_key_sel_o != mr_round_key_sel[i]) |-> mr_err == 1'b1 ;  
 endproperty 
\end{lstlisting}
\end{mybox}

%% file: Appendix_exmaples_Addressing_Issues/HumanExpert.tex
\begin{mybox}[colback=white, colframe=black, boxrule=0.2mm, sharp corners, coltitle=black, colbacktitle=white, label=tbox:hum_exp, breakable] {Examples comparing LLMs Fine-tuned on VERT to Human Experts.}
\lstset{style=mystyle2}

\textbf{Example 1}

Source Code:
\begin{lstlisting}[language=Verilog]
if (!rst_ni) begin
            	bus_release_cnt <= '0;  
end else if (monitor_enable &&!monitor_enable_q) begin
            	if (multi_controller_enable_i) begin
            	bus_release_cnt <= 30'(bus_inactive_timeout_i);
            	end
end else if (bus_release_cnt_load) begin
            	bus_release_cnt <= bus_release_cnt_sel;
end else if (bus_release_cnt_dec && (bus_release_cnt!= '0)) begin
            	bus_release_cnt <= bus_release_cnt - 1'b1;
end

\end{lstlisting}

Llama 3.1 fine-tuned on \dataset Response:

\begin{lstlisting}[language=Verilog]
 property ResetSynceotid;
  ( !rst_ni ) |-> bus_release_cnt == '0 ;
  endproperty
 property ResetSynceotid;
  (  rst_ni ) && ( !monitor_enable || monitor_enable_q ) |-> bus_release_cnt == '0 ;
  endproperty
property ResetSynceotid;
  (  rst_ni ) &&  ( monitor_enable && !monitor_enable_q ) &&  (multi_controller_enable_i ) |-> bus_release_cnt == 30'(bus_inactive_timeout_i);
  endproperty
property ResetSynceotid;
  (  rst_ni ) &&  ( !monitor_enable || monitor_enable_q ) && (  bus_release_cnt_load ) |-> bus_release_cnt == bus_release_cnt_sel ;
  endproperty
property ResetSynceotid;
  (  rst_ni ) &&  ( !monitor_enable || monitor_enable_q ) && ( !bus_release_cnt_load ) &&  (  bus_release_cnt_dec && (bus_release_cnt != '0)) |-> bus_release_cnt == (bus_release_cnt - 1'b1) ;
  endproperty
\end{lstlisting}

DeepSeek Coder fine-tuned on \dataset  Response:

\begin{lstlisting}[language=Verilog]
property ResetSynceotid;
  ( !rst_ni ) |-> bus_release_cnt == '0;
  endproperty
property ResetSynceotid;
  (  rst_ni ) && ( !monitor_enable || monitor_enable_q ) |-> bus_release_cnt == '0;
  endproperty
property ResetSynceotid;
  (  rst_ni ) &&  ( monitor_enable && !monitor_enable_q ) &&  (  multi_controller_enable_i ) |-> bus_release_cnt == 30'(bus_inactive_timeout_i) ;
  endproperty
property ResetSynceotid;
  (  rst_ni ) &&  ( !monitor_enable || monitor_enable_q ) && (  bus_release_cnt_load ) |-> bus_release_cnt == bus_release_cnt_sel ;
  endproperty
property ResetSynceotid;
  (  rst_ni ) &&  ( !monitor_enable || monitor_enable_q ) && ( !bus_release_cnt_load ) &&  (  bus_release_cnt_dec && (bus_release_cnt != '0)) |-> bus_release_cnt == (bus_release_cnt - 1'b1) ;
  endproperty
\end{lstlisting}

Human Expert:

\begin{lstlisting}[language=Verilog]
property bus_release_cnt_change_reset;
  !rst_ni |-> !(|bus_release_cnt);
endproperty
property bus_release_cnt_change;
  rst_ni && monitor_enable && !monitor_enable_q |-> multi_controller_enable_i |-> bus_release_cnt == 30'(bus_inactive_timeout_i);
endproperty
property bus_release_cnt_change_load;
  rst_ni && bus_release_cnt_load |-> bus_release_cnt == bus_release_cnt_sel;
endproperty
property bus_release_cnt_change_dec;
  rst_ni&&bus_release_cnt_dec&&(|bus_release_cnt)|->bus_release_cnt==bus_release_cnt-1;
endproperty
\end{lstlisting}

\hspace*{-0.5cm}\rule{1.078\textwidth}{0.4pt}









\textbf{Example 2}

Source Code:
\begin{lstlisting}[language=Verilog]
    if (state_sel_o            	!= mr_state_sel[i]     	||
    add_rk_sel_o  != mr_add_rk_sel[i] 	||
    key_full_sel_o  != mr_key_full_sel[i]  ||
    key_dec_sel_o   != mr_key_dec_sel[i]   ||
    key_words_sel_o != mr_key_words_sel[i] ||
    round_key_sel_o != mr_round_key_sel[i]) begin
        mr_err = 1'b1;
    end

\end{lstlisting}

Llama 3.1 fine-tuned on \dataset Response:

\begin{lstlisting}[language=Verilog]
property SyncErreotid;
  (state_sel_o	!= mr_state_sel[i]     	||
  add_rk_sel_o   != mr_add_rk_sel[i]  ||
  key_full_sel_o != mr_key_full_sel[i]  ||
  key_dec_sel_o  != mr_key_dec_sel[i]   ||
  key_words_sel_o!= mr_key_words_sel[i] ||
  round_key_sel_o!= mr_round_key_sel[i]) |-> mr_err = 1'b1;
endproperty
\end{lstlisting}

DeepSeek Coder fine-tuned on \dataset Response:

\begin{lstlisting}[language=Verilog]
property SyncErreotid;
  (state_sel_o   != mr_state_sel[i]) ||
  (add_rk_sel_o           	!= mr_add_rk_sel[i]) ||
  (key_full_sel_o  != mr_key_full_sel[i]) ||
  (key_dec_sel_o   != mr_key_dec_sel[i]) ||
  (key_words_sel_o != mr_key_words_sel[i]) ||
  (round_key_sel_o != mr_round_key_sel[i]) |-> mr_err == 1'b1 ;
endproperty
\end{lstlisting}

Human Expert:

\begin{lstlisting}[language=Verilog]
property state_sel_cond;
  (state_sel_o != mr_state_sel[i]) || (add_rk_sel_o != mr_add_rk_sel[i]) || (key_full_sel_o != mr_key_full_sel[i]) || (key_dec_sel_o != mr_key_dec_sel[i]) || (key_words_sel_o != mr_key_words_sel[i]) || (round_key_sel_o != mr_round_key_sel[i]) |-> mr_err;
endproperty
\end{lstlisting}

\end{mybox}

%% file: Appendix_syn_func/Keypoint_quail.tex
\begin{fncyBox}
    \textbf{Key Point:} Qualitative comparisons reveal that LLMs fine-tuned on VERT overcome common pitfalls (such as misinterpreting clock cycles and nested conditions) to produce assertions that closely align with human expert quality, reinforcing the dataset’s effectiveness in complex verification scenarios.
\end{fncyBox}

%% file: Tables/contamination.tex
\begin{table}[b]
\centering
\caption{Effect of increasing contamination in design files on assertion generation}
\label{tab:cont-table}
\renewcommand{\arraystretch}{1.5}
\resizebox{\textwidth}{!}{%
\begin{tabular}{c|c|ccc|ccc|ccc}
\hline
\multirow{2}{*}{\textbf{Models}} &
  \multirow{2}{*}{\textbf{Benchmark/Hardware IP}} &
  \multicolumn{3}{c|}{\textbf{Generated Assertions}} &
  \multicolumn{3}{c|}{\textbf{Syntactically Correct (\%)}} &
  \multicolumn{3}{c}{\textbf{Functionally Correct (\%)}} \\
 &
   &
  \multicolumn{1}{c|}{\textbf{No Contamination}} &
  \multicolumn{1}{c|}{\textbf{+10 Contamination}} &
  \textbf{+20 Contamination} &
  \multicolumn{1}{c|}{\textbf{No Contamination}} &
  \multicolumn{1}{c|}{\textbf{+10 Contamination}} &
  \textbf{+20 Contamination} &
  \multicolumn{1}{c|}{\textbf{No Contamination}} &
  \multicolumn{1}{c|}{\textbf{+10 Contamination}} &
  \textbf{+20 Contamination} \\ \hline
\multirow{16}{*}{\textbf{Llama 3.1}}      & OpenTitan/AES          & 125 & 129 & 132 & 89 & 86 & 84 & 83 & 81 & 79 \\
                                          & OpenTitan/I2C          & 126 & 130 & 132 & 83 & 81 & 80  & 83 & 81 & 80  \\
                                          & OpenTitan/LC CTRL      & 19  & 21  & 22  & 89 & 81 & 77 & 89 & 81 & 77 \\
                                          & OpenTitan/ADC CTRL     & 32  & 34  & 36  & 100 & 94 & 89 & 100 & 94 & 89 \\ \cline{2-11} 
                                          & CVA6/Frontend          & 13  & 14  & 17  & 92 & 86 & 71 & 92 & 86 & 71 \\
                                          & CVA6/Decode\&Issue     & 34  & 37  & 39  & 100 & 92 & 87 & 100 & 92 & 87 \\
                                          & CVA6/Execute           & 105 & 109 & 111 & 91 & 88 & 86 & 91 & 88 & 86 \\
                                          & CVA6/Commit            & 79  & 82  & 84  & 90 & 87 & 85 & 90 & 87 & 85 \\
                                          & CVA6/Controller\&Top   & 68  & 71  & 73  & 96 & 92 & 89 & 96 & 92 & 89 \\ \cline{2-11} 
                                          & Pulpissimo/APB         & 19  & 21  & 23  & 89 & 81 & 74 & 89 & 81 & 74 \\
                                          & Pulpissimo/RISCV       & 15  & 17  & 18  & 93 & 82 & 78 & 93 & 82 & 78 \\
                                          & Pulpissimo/debug\_unit & 11  & 14  & 14  & 100 & 93 & 79 & 100 & 83 & 81 \\ \cline{2-11}
                                          & OpenPiton/IO\_CTRL     & 97  & 102  & 107  & 99 & 96 & 88 & 99 & 91 & 86 \\
                                          & OpenPiton/JTAG         & 36  & 39  & 42  & 96 & 92 & 86 & 96 & 88 & 83\\
                                          & OpenPiton/MEM\_IO      & 20  & 23  & 27  & 100 & 94 & 81 & 100 & 93 & 89\\
                                          & OpenPiton/NOC\_BRIDGE  & 24  & 27  & 29  & 98 & 91 & 89 & 98 & 89 & 85 \\ 
                                          \hline
\multirow{16}{*}{\textbf{DeepSeek Coder}} & OpenTitan/AES          & 157 & 161 & 164 & 95 & 93 & 91 & 94 & 91 & 90  \\
                                          & OpenTitan/I2C          & 124 & 129 & 131 & 98 & 94 & 92 & 98 & 94 & 92 \\
                                          & OpenTitan/LC CTRL      & 19  & 22  & 23  & 100 & 86 & 83 & 100 & 86 & 83 \\
                                          & OpenTitan/ADC CTRL     & 32  & 35  & 35  & 100 & 91 & 91 & 97 & 89 & 89 \\ \cline{2-11} 
                                          & CVA6/Frontend          & 14  & 16  & 18  & 93 & 81 & 72 & 93 & 81 & 72 \\
                                          & CVA6/Decode\&Issue     & 32  & 34  & 35  & 100 & 94 & 91 & 100 & 94 & 91 \\
                                          & CVA6/Execute           & 99  & 102 & 104 & 98 & 95 & 93 & 98 & 95 & 93 \\
                                          & CVA6/Commit            & 93  & 95  & 96  & 89 & 87 & 86 & 89 & 87 & 86 \\
                                          & CVA6/Controller\&Top   & 76  & 81  & 81  & 89 & 84 & 84 & 89 & 84 & 84 \\ \cline{2-11} 
                                          & Pulpissimo/APB         & 19  & 22  & 22  & 100 & 86 & 86 & 100 & 86 & 86 \\
                                          & Pulpissimo/RISCV       & 15  & 18  & 19  & 100 & 83 & 79 & 100 & 83 & 79 \\
                                          & Pulpissimo/debug\_unit & 11  & 13  & 15  & 100 & 85 & 73 & 100 & 85 & 73 \\ \cline{2-11}
                                          & OpenPiton/IO\_CTRL     & 103 & 107  & 111 & 92 & 90 & 89 & 86 & 85 & 84 \\
                                          & OpenPiton/JTAG         & 27  & 31  & 36  & 100 & 88 & 81 & 89 & 81 & 77\\
                                          & OpenPiton/MEM\_IO      & 72  & 76  & 79  & 96 & 91 & 86 & 93 & 91 & 85\\
                                          & OpenPiton/NOC\_BRIDGE  & 35  &  39  & 43  & 94 & 91 & 85 & 94 & 87 & 77 \\ 
                                          \hline
                                    
\end{tabular}%
}
\end{table}

%% file: sections/conclusion.tex
\section{Conclusion}
\label{conclusion}

In this paper, we introduce \dataset, a novel open-source dataset tailored to automate the generation of SystemVerilog assertions, enabling a more scalable and efficient hardware verification process using LLMs. By systematically fine-tuning popular models such as DeepSeek Coder and LLaMA 3.1 on our dataset, we achieved substantial improvements in both syntactical accuracy and functional correctness of generated assertions across real-world SoCs, including OpenTitan, CVA6, Pulpissimo, and OpenPiton. Our evaluation demonstrated the adaptability of these LLMs, fine-tuned with VERT, furnishing up to a 96.88\% improvement in both functional and syntactical correctness over base models and up to 24.14\% over GPT-4o. This work is the first to demonstrate the potential of combining domain-specific datasets with advanced LLMs to address the enhanced challenges of modern hardware verification. In the future, we will focus on expanding the dataset to cover more intricate design patterns and hardware architectures, as well as improving model performance in handling asynchronous and synchronous conditions. Moreover, we aim to integrate our approach with industry-standard functional verification tools to streamline the hardware verification process. 